\documentclass[twoside]{article}

\usepackage[accepted]{aistats2022}
\usepackage{graphicx}
\usepackage[none]{hyphenat}

\usepackage[round]{natbib}

\usepackage{graphicx}
\usepackage{amsmath,amsfonts,amssymb}
\usepackage{bm}
\usepackage{nicefrac}
\usepackage{booktabs}
\usepackage{diagbox}
\usepackage{soul,color}
\usepackage{subfig}
\usepackage{wrapfig}
\usepackage{multirow}

\DeclareSymbolFont{symbolsC}{U}{pxsyc}{m}{n}
\DeclareMathSymbol{\coloneqq}{\mathrel}{symbolsC}{"42}
\newcommand{\R}{\mathbb{R}}

\begin{document}



\twocolumn[

%\aistatstitle{Equivariant Representation Learning for Deep Dynamical Modeling}
\aistatstitle{Equivariant Deep Dynamical Model for Motion Prediction}
\aistatsauthor{ Bahar Azari \And Deniz Erdo\u{g}mu\c{s}}

\aistatsaddress{ Northeastern University\\ 
            \texttt{azari.b@northeastern.edu} \And
            Northeastern University\\
            \texttt{Erdogmus@ece.neu.edu}} ]

\begin{abstract}
Learning representations through deep generative modeling is a powerful approach for dynamical modeling to discover the most simplified and compressed underlying description of the data, to then use it for other tasks such as prediction. Most learning tasks have intrinsic symmetries, i.e., the input transformations leave the output unchanged, or the output undergoes a similar transformation. The learning process is, however, usually uninformed of these symmetries. Therefore, the learned representations for individually transformed inputs may not be meaningfully related. In this paper, we propose an SO(3) equivariant deep dynamical model (EqDDM) for motion prediction that learns a structured representation of the input space in the sense that the embedding varies with symmetry transformations. EqDDM is equipped with equivariant networks to parameterize the state-space emission and transition models.  We demonstrate the superior predictive performance of the proposed model on various motion data.  
\end{abstract}

\section{Introduction}
\label{sec:intro}
\begin{figure}[!t]
    \centering
    \begin{minipage}[l]{0.53\linewidth}
    \caption{Rotation is one of the symmetries of the swinging pendulum as a dynamic system. Rotations of the coordinate system, although alter the relative position of the points in the collected data, leave the underlying dynamical system unchanged.\label{fig:pen}}
    \end{minipage}\begin{minipage}[r]{0.47\linewidth}
    \hfill\includegraphics[width=.95\linewidth]{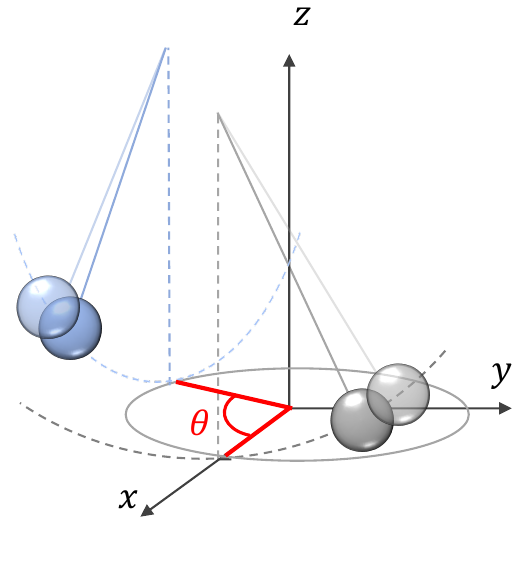}
    \end{minipage}
\end{figure}

Deep dynamical system models are introduced to cope with the potential non-linearities found in real-world systems. These models are constructed in such a way that they learn rich but compressed representations of the given data through structured deep generative modeling that captures the underlying complex distribution of data \citep{watter2015embed,karl2017deep,krishnan2017structured,linderman2017bayesian,fraccaro2017disentangled,becker2019recurrent,nassar2019tree,farnoosh2021dynamical, farnoosh2021deep,shamsabardeh2021prrs}. The input to the models is usually multiple realizations of the system behavior we want to capture, called trajectories. Each realization is a set of spatially correlated time series consisting of measurements such as 3D coordinates indicative of the position of an object collected over time, such as the position of joints in skeletal data or the ball coordinates in swinging pendulum. Inevitably, the predictive model that is trained on these trajectories cannot be used for another one where the relative position and orientation of the coordinates system have changed. Figure.~\ref{fig:pen} shows two otherwise similar pendulums placed in two planes parallel to $z$-axis and rotated by $\theta$ with respect to one another. It is intuitive to see that the underlying dynamic of the swinging pendulum, i.e., its angular acceleration and velocity, remains the same regardless of its relative orientation to the coordinate system. This is because rotation is one of the symmetries of the swinging pendulum as a dynamic system. One of the main limitations of the current generative models is that they do not consider the symmetries of the model at hand.

\par Symmetry refers to a transformation that leaves an object (or its higher-level representation) invariant. Symmetries can also be associated with tasks. For example, translations are symmetries of the object classification task, and so where the object is inside an image should not matter to the classifier. Therefore, a model (e.g., a neural network) should process different but correspondent versions of an object under these transformations equivalently. Exploiting symmetry has a long history in physical sciences. However, recently, many studies in the literature focused on incorporating symmetries into variants of deep neural networks to learn images, sets, point clouds, and graphs,  (\citealt{cohen2016steerable, kondor2018generalization, maron2018invariant, cohen2019general, NEURIPS2019_ea9268cb, feige2019invariant,  wang2021incorporating, walters2021trajectory}).

\par In most of the studies, with the help of mathematical tools such as group theory and representation theory, a \emph{global} architecture is designed for a learning problem in such a way that it is invariant or equivariant under various transformations of the input. In other words, the input is treated as a single object whose symmetries should be preserved throughout the network. However, to the best of our knowledge, limited works exist investigating the integration of equivariance/invariance into the more complex Markovian generative models. The complicated structure of these models, which are associated with multiple levels of latent variables connected through neural networks,  renders the design of an efficient equivariant model more challenging. Specifically, in this scenario, we need to consider each \emph{local} network connecting a part of an input to its latent, or a current latent to the future latent, etc.

\par In this paper, we investigate the role of symmetry in learning of a dynamic system characterized with an equivariant deep state space model. We propose an $\textup{SO}(3)$-equivariance deep dynamical model for motion prediction, and call it EqDDM. Our model, which inherits a Markovian structure from its state-space model counterpart, is equipped with a chain of appropriately equivariant/invariant MLPs similar to those in \cite{finzi21a}. Specifically, we propose to use a hierarchical equivariant structure where we have an equivariant network from each input to its representation (latent), followed by an equivariant switching network from the current latent to the future latent, followed by an invariant switching network controlling each switch (see figure.~\ref{fig:pgm} \textbf{Right}).

\par The structure of the paper is organized as follows. After covering the relevant studies around the subject in section.~\ref{sec:related}, we review the necessary background on group theory, representation theory, and state-space models in section.~\ref{sec:back}. We then outline the designing steps of an equivariant/invariant network in section.~\ref{sec:method}. In section.~\ref{sec:EqDDM}, we characterize our dynamic learning problem  with a generative state space model. We then describe how to integrate equivariant/invariant architectures in a Markovian structure.  We evaluate the performance of our model on motion prediction of a skeletal object and provide the results in section.~\ref{sec:res}.

%\textbf{Equivariance preserve symmetry.(Later layer can exploit the symmetry) It is because the output feature map also undergoes a translation when the input does, that a pattern in the output feature map can also occur at any location stack convolution layer.}

\section{Related Work}
\label{sec:related}
Incorporating symmetry into deep neural networks has been the focus in many studies due to its compelling promise to improve generalization and accuracy.  The line of research began with the attempt to generalize the idea of the translation equivariant convolution layer in CNN to other spaces and symmetry groups, as one can associate the ability of CNN with the fact that it exploits the translational symmetry \citep{gens2014deep, olah2014groups,dieleman2015rotation,guttenberg2016permutation,dieleman2016exploiting,cohen2016steerable, ravanbakhsh2016deep,ravanbakhsh2017equivariance,worrall2017harmonic,maron2020learning,dym2020universality, finzi20a,satorras2021n}. Early studies focused on discrete groups for their ease of understanding \citep{cohen2016group, maron2018invariant,zaheer2017deep}. Some works have been investigating equivariance to continuous groups and generalized the CNN to various spaces  \citep{cohen2018spherical, kondor2018generalization, cohen2019general, walters2021trajectory, azari2021circular}. Lately, \cite{finzi21a} has has generalized quivariant multilayer perceptrons (MLPs) for arbitrary matrix groups. 

Incorporating symmetry into complex probabilistic deep generative models has not been entirely examined. Some studies tackled the problem of learning equivariant and invariant representations using variational autoencoders  \citep{feige2019invariant, esteves2018learning, qi2019avt,gao2020graphter,kohler2020equivariant}. Other studies integrated symmetries in dynamic models by defining a new equivariant convolutional layer \citep{walters2021trajectory,wang2021incorporating}. To the best of our knowledge, our study is the first attempt at designing an equivariant deep state-space model.

\section{Background}
\label{sec:back}
We begin by explaining the required building blocks of our model starting from Lie groups and their representations. Note that in this paper, we use the word representation in two different contexts: representation as a compact but informative numeric feature that captures relevant information regarding an input signal, and representation as an invertible matrix, representative of a group element, that act on a vector space.

\subsection{Lie Group \& Infinitesimal Generator}
\begin{figure}[!t]
%\vspace{.3in}
\centerline{\includegraphics[width=.80\linewidth]{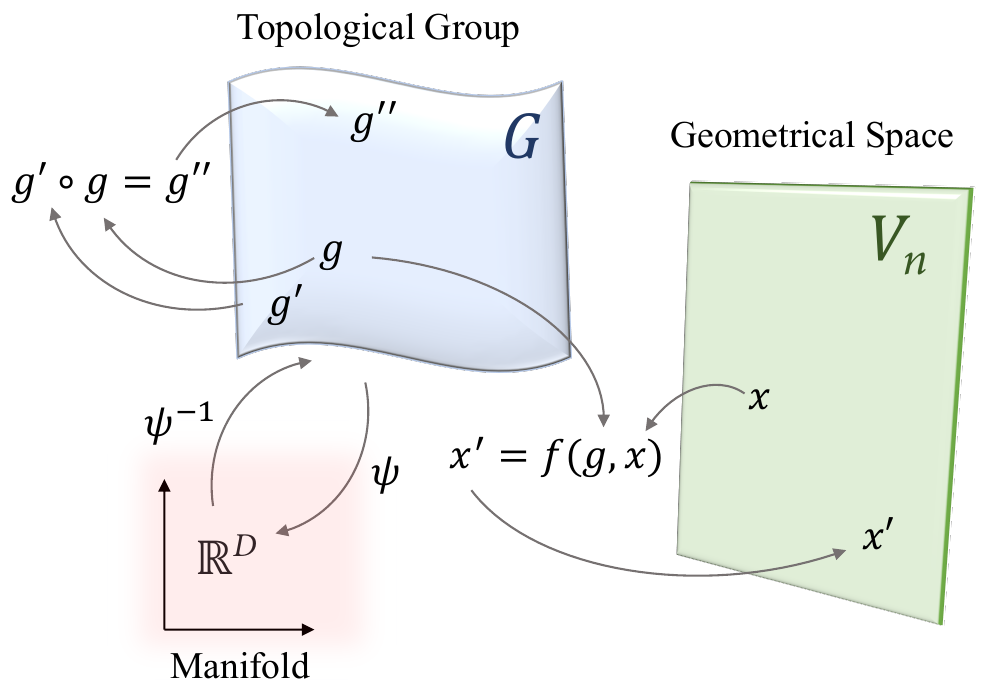}}
%\vspace{.3in}
\caption{A Lie group of $n\times n$ matrices and its corresponding $D$ dimensional manifold acting on a $n$ dimensional vector space $V_n$. 
}
\label{fig:topo}
\end{figure}

A Lie group $G$ is a smooth manifold equipped with the structure of a group such that the group operation and inverse-assigning operation are smooth functions. The manifold is locally represented by a chart mapping ($\psi$) to an underlying Euclidean space $\mathbb{R}^D$, where $D$ is the dimensionality of the manifold.  Furthermore, the chart map is defined in such a way that it associates the identity element in the group with the origin of Euclidean space. Elements of the Lie group can \textit{act} as a transformation on the basis an $n$-dimensional vector space known as the \textit{geometric space}, and change the coordinates of elements accordingly (see figure.~\ref{fig:topo} for a visualisation). We analyze Lie groups in terms of their infinitesimal generators which are the derivative of the group elements with respect to its $D$ underlying parameters at the identity. These infinitesimal generators are $n\times n$ matrices $\{A_1, A_2,\dots, A_D\}$, that are the bases for a new vector space, or more strongly an Algebra, called the Lie Algebra $\mathfrak{g}$.  We can relate an element in the Lie Algebra, which is a linear combination of the generator matrices, to an element in the Lie group through the exponential map $\exp: \mathfrak{g} \to G$ as $g =\exp{(\sum_{i=1}^D\alpha_iA_i)}$ (refer to \cite{gilmore2006lie} for more details).

\subsection{Lie Algebra Representations} A linear finite dimensional group representation is a map $\rho:G\to\text{GL}_n(\mathbb{R})$ from the abstract notion of a group element to an invertible matrix belonging to the general linear group $\text{GL}_n(\mathbb{R})$ -- the set of $n\times n$ invertible matrices. The group element represented in this form is the transformation that acts on the elements (i.e., their corresponding coordinates) in a vector space $V_n$ called the geometric space.    The representation should satisfy $\rho(g_1 g_2) = \rho(g_1 g_2) = \rho(g_1)\rho(g_2)$ and consequently $\rho(g^{-1}) = \rho(g)^{-1}$ for all $g_1, g_2 \in G$. Furthermore, Lie Algebra of a Lie group has a corresponding matrix representation denoted as $d\rho: \mathfrak{g} \to \mathfrak{gl}_n(\mathbb{R})$ that linearly maps Lie Algebra elements to
the set of  $n \times n$ matrices (commonly denoted as $\mathfrak{gl}_n(\mathbb{R})$). The representation of a Lie Group and the that of its
Lie Algebra is related by:
\begin{equation}
\rho(e^{A}) = e^{d\rho(A)} \qquad \forall A\in \mathfrak{g}.
\label{eqn:Lie}
\end{equation}
\subsection{Lie group $\textup{SO}(n)$}
For the rotation group $\textup{SO}(n)$, the dimensionality of the group and its Lie Algebra, $D=\text{dim}(\mathfrak{so}(n)) = \text{dim}(\textup{SO}(n))$, is given by  $\nicefrac{n(n-1)}{2}$. The group elements are
matrices $R\in \mathbb{R}^{n\times n}$ such that $R^\top R = I$ and $\det(R) = 1$, and the anti-symmetric matrices constitute its Lie Algebra $\mathfrak{so}(n) = \{A\in\mathbb{R}^{n\times n}:A^\top = -A\}$.
%constitute the basis of a vector space, or more strongly an Algebra, called the Lie Algebra, and so their linear combination is used to create a group element 
\subsection{Tensor Representations}Tensor manipulation is required to build increasingly complex representation matrices with increasingly more parameters to construct an expressive neural network function. Given a base group representation $\rho$ acting on a vector space $V$, and its corresponding Lie Algebra representation $d\rho$,  we can design larger and more sophisticated representations using the following tensor operations: 
Direct sum ($\oplus$) acts on matrices and concatenates them on the diagonal as $X\oplus Y = \begin{bmatrix}
X & 0 \\
0 & Y 
\end{bmatrix}$. Tensor Product($\otimes$) is the standard Kronecker product. Finally, $V^{\ast}$  is the dual space of $V$, and its corresponding Lie group and Lie Algebra representation are $\rho(g^{-1})^{\top}$ and $-d\rho(A)^{\top}$, respectively. Using the tensor product and dual operator, we can describe linear maps between two vector spaces. Linear maps from $V_1 \to V_2$ form the vector space $V_2 \otimes V_1^{\ast}$ and have the corresponding representation $\rho_2\otimes\rho_1^{\ast}$. More notations used throughout the paper are as follows. We denote several copies of the same vector space
$\underbrace{V\oplus V\oplus \dots \oplus V}_{m}$ as $mV$. We also refer to the vector space formed by several tensor products as $T_{(p,q)} = V^{\otimes p} \otimes V^{\ast\otimes q}$ where $(\cdot)^{\otimes p}$is the tensor product iterated
$p$ times. 

%\vspace{-.3cm}

\section{$\textup{SO}\mathbf{(3)}$-equivariant Network Design}
\label{sec:method}
In this section, we describe the general framework for designing an equivariant (invariant) architecture inspired by \cite{finzi21a}, to then deploy it in our deep generative model with Markovian structure. Specifically, we focus on designing an $\textup{SO}(3)$-equivariant network since $\textup{SO}(3)$ constitutes the symmetries of the representation learning in the dynamic modeling problem at hand. We start with a general Lie group $G$ and then provide the specific results for $\textup{SO}(3)$. 

\subsection{Equivariant Linear Layer}
We require the group transformation that the input undergoes to traverse the network and appear in the output (e.g., latent space). This goal is achievable by defining each layer of the network to be equivariant to the group action. To formally establish the equivariance property, we define the vector spaces $V_1$ and $V_2$ (of dimensionality $N_1$ and $N_2$) to represent the input and output of a network layer, respectively. We also define the action of the group on these two vector spaces as transformation matrices $\rho_1(g): G \to \textup{GL}_{N_1}(\R)$ and $\rho_2(g): G \to \textup{GL}_{N_2}(\R)$. An equivariant linear layer is parameterized with the weight matrix $W\in \R^{N_2\times N_1}$ that maps the $V_1$ to $V_2$. Equivariance implies that transforming input is equivalent to correspondingly transforming the output and since it is true for all the input $x \in V_1$, we can say:
\begin{equation}
    \rho_2(g)W = W\rho_1(g) \qquad \forall g \in G.
    \nonumber
\end{equation}
We can simplify the equality by using the tensor product and $\mathit{vec}(\cdot)$ operator manipulation (see the supplementary for details) to reach to:
\begin{equation}
\left(\rho_2(g)\otimes\rho_1(g^{-1})^\top\right)\mathit{vec}(W) = \mathit{vec}(W) \quad \forall g \in G, \label{eqn:equi}
\end{equation} where the $\mathit{vec}(\cdot)$ operator creates a column vector from a matrix by stacking the its column vectors below one another. The representation $\rho_1(g^{-1})^\top$ is the dual representation $\rho_1^{\ast}(g)$. If we closely examine equation.~\ref{eqn:equi}, we realize that it is similar to an invariance equality where the representation $\left(\rho_2\otimes\rho_1^{\ast}\right)(g) = \rho_2(g)\otimes\rho_1(g^{-1})^\top$ on the left-hand side is just a more intricate representation composed of two simple ones through the tensor operations. Specifically, we can say the vectorized version of matrices mapping from $V_1 \to V_2$ create the vector space $V_2 \otimes V_1^{\ast}$, and every group element $g$ acts on it with the representation $\rho = \left(\rho_2\otimes\rho_1^{\ast}\right)$:
\begin{equation}
\rho(g)v = v \quad \forall g \in G,\; \forall v \in V_2 \otimes V_1^{\ast}. \label{eqn:rep}
\end{equation}
Elements in $G$ can be finitely generated by taking the $\exp(\cdot)$ of some linear combination of its Lie Algebra bases (the generators). Hence, we can write the representation of $g$ in terms of the exponential of the generator matrices, $\{A_i\}_{i=1}^D$, and relate it to its Lie algebra representation  as: 
\begin{equation}
\rho(g)=\rho(\exp{(\sum_{i=1}^D\alpha_i A_i)}) = \exp{(\sum_{i=1}^D\alpha_i d\rho(A_i)}) \quad \forall \alpha_i.
\end{equation}
Note that for the latter equality, we used the correspondence in (\ref{eqn:Lie}) and the fact that $d\rho(\cdot)$ is linear. Substituting the obtained $\rho(g)$ in (\ref{eqn:rep}) we have:\begin{equation}
\exp{(\sum_{i=1}^D\alpha_i d\rho(A_i)})v = v \quad \forall \alpha_i.
\label{eqn:rep2}
\end{equation}
Since (\ref{eqn:rep2}) is true for all $\alpha_i$s, it is also true for its derivative with respect to $\alpha_i$ at $\alpha= 0$. Hence, we get $D$ constraints collected in a larger matrix as:
\begin{equation}
Cv=
\begin{bmatrix}
d\rho(A_1)\\
\vdots\\
d\rho(A_D)
\end{bmatrix}v=0.
\label{eqn:const}
\end{equation}Note that since all the $A_i$s are known, and different $\rho(A_i)$ can be constructed using the tensor representation, the problem in (\ref{eqn:const}) is a standard nullspace problem addressed in Finzi et al. Specifically, they used a Krylov method for efficiently solving the nullspace problem by exploiting structure in the matrices $\rho$ and $d\rho$ (see \cite{finzi21a} for more detail). The obtained nullspace $Q\in\mathbb{R}^{n\times r}$ can apply symmetry to the arbitrary weight matrix flattened as $v$, where $r$ is the rank of the nullspace, and $n$ is the dimensionality of $v$.  In practice, we can parameterize a set of weights $v_0$ and project them onto the equivariant subspace by $v=QQ^\top v_0$, which encourages the weight-sharing scheme. Through the linear combination induced by the projection, the unique elements in $v_0$ are projected into a set of repeating (i.e., \emph{shared}) elements in $v$. Although this structured weight sharing promotes equivariance, it comes with the caveat that the expressive power of the network is restricted in that the number of parameters is now limited. For example, for $G=\textup{SO}(3)$, the weight matrix $W$ reshaped from rows of $v$ obtained from solving (\ref{eqn:const}) has only one non-zero parameter, and hence limited expressive power.  
  
\par To remedy this issue, we need to use more complex representations, constructed through tensor operations, to obtain more parameters through achieving a higher nullspace rank. Considering a feature space $U$ in a neural network, it can be a combination of tensors with different ranks. For example, if we consider the group $\textup{SO}(3)$ acting on $\mathbb{R}^3$, a complex feature space can be constructed using scalars, denoted as $T_0$, $3$ dimensional vectors, denoted as $T_1$, $3\times 3$ matrix denoted as $T_2$, and so on. Using the notation described in the background section, we can write $U = c_0T_0 \oplus c_1T_1 \oplus \, \dots \oplus c_MT_M$. The representation for a linear map $U_1 \to U2$ between two feature space in the network with the corresponding representations $\rho_{U_1}(g) = \bigoplus_{a\in\mathcal{A}_1} \rho_a(g)$ and $\rho_{U_2}(g) = \bigoplus_{b\in\mathcal{A}_2} \rho_b(g)$, is given by:
\begin{equation}
    \rho_2 \otimes \rho_1^{\ast} =   \bigoplus_{b\in\mathcal{A}_2} \rho_b \otimes \bigoplus_{a\in\mathcal{A}_1} \rho_a^{\ast} = \bigoplus_{(b,a)\in\mathcal{A}_2\times \mathcal{A}_1} \rho_b\otimes \rho_a^{\ast}.
\end{equation}
To design a fully equivariant MLP from the described equivariant linear layer, we use the gated nonlinearities introduced in \citep{weiler20183d}, and used in \cite{finzi21a} as an equivariant nonlinearity. In the next section, we propose our equivariant deep dynamic model using the building blocks described here.

\section{Equivariant Deep Dynamical Model}
\label{sec:EqDDM}
We propose an equivariant deep dynamical model (EqDDM) based on the equivariant  linear map described in the section.~\ref{sec:method}. We describe its associated  generative and inference model in the following. 

\begin{figure*}[!t]
\centering
\includegraphics[width=.9\linewidth]{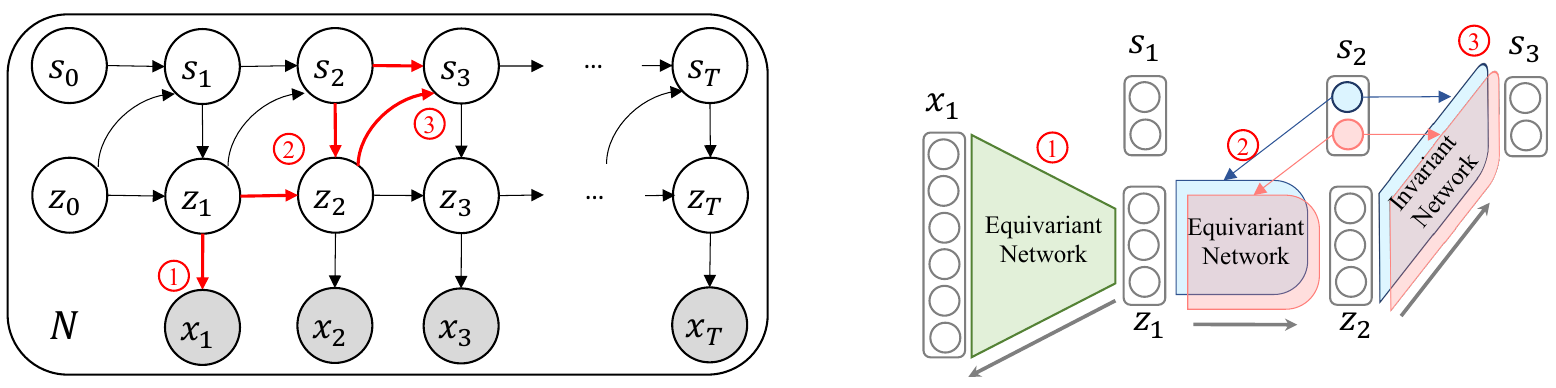}
\caption{\textbf{Left:} Probabilistic graphical model of equivariant deep state space model. \textbf{Right:} The generative equivariant and invariant networks are for edges in red. To avoid crowding the figure, we only draw some of the networks in the generative model on the left, but the pattern repeats throughout the chain.  The label (1) is the equivariant generative network for constructing $x_t$ using the latent $z_t$, label (2) is the equivariant generative network informing the current latent $z_{t}$ of the past latent $z_{t-1}$ governed by the current state $s_{t}$, and label (3) is the invariant network influencing future state  $s_{t+1}$ using the current latent  $z_{t}$ and state $s_{t}$. Arrows show the direction of the forward pass of the networks.}
\label{fig:pgm}
\end{figure*}
\subsection{Generative Model}
\label{sec:prob}

Lets consider a set of $N$ motion sequences $\{X_1, \dots, X_N\}$, where each sequence $X_n \in \mathbb{R}^{T_n \times (D\times3)}$ records 3D coordinates of $D$ objects/skeletal joints over $T_n$ time points. A switching dynamical model defines a generative distribution over this dataset according to a set of \emph{discrete} dynamical states $\mathcal{S}_n=\{s_{n,t}\}_{t=1}^T$ and their corresponding \emph{continuous} temporal latents $Z_n = \{z_{n,t}\in \mathbb{R}^{K}\}_{t=1}^T$ as follows:
\begin{align}
    x_{n,t} &\sim p_\theta(x_{n,t}\,|\,z_{n,t}),\nonumber\\
    z_{n,t} &\sim p_\theta(z_{n,t}\,|\,z_{n,t-\ell},s_{n,t}),\nonumber\\
    s_{n,t} &\sim p_\theta(s_{n,t}\,|\,s_{n,t-1},z_{n,t-1}),
    \label{eqn:pgen}
\end{align}
where $\theta$ collectively denotes generative distribution parameters and $\ell$ indicates a set of temporal lags (e.g., $\ell=\{1,2\}$ for a second-order model). The probabilistic graphical model for EqDDM is shown in \ref{fig:pgm}.    

Specifically, the distributions $p_\theta(s_{n,t}|s_{n,t-1},z_{n,t-1})$ define a discrete Markovian prior over $\mathcal{S}_n$ (subscript $n$ is dropped hereafter for brevity):
\begin{align}
    p_\theta(s_{t}\,|\,s_{t-1}=s, z_{t-1}) = \text{Cat}\left(\boldsymbol{\pi_\theta}^s\left(z_{t-1}\right)\right),
\end{align}
where $\boldsymbol{\pi}_{\boldsymbol{\theta}}^s(\cdot)$ is a state transition network that is set by the preceding state $s_{t-1}$ and maps $z_{t-1}$ to the prior distribution parameters of $s_t$ (a.k.a. a recurrent state transition model \citep{linderman2017bayesian}). This probability is characterized by an invariant switching network. Note that the state of the system should be invariant to the translations of the latent variables.   

The distributions $p_\theta(z_{n,t}\,|\,z_{n,t-\ell},s_{n,t})$ define a switching dynamical autoregressive prior over $Z_n$ (a.k.a. a transition model): 
\begin{align}
   p_\theta(z_{t}|z_{t-\ell},s_{t} = s) =
   \text{Norm}\Big(\boldsymbol{\mu_{\theta}}^s(z_{t-\ell}),
   \boldsymbol{\sigma_{\theta}}^s(z_{t-\ell})\Big),
   \label{eqn:trans}
\end{align}
where state-specific $\boldsymbol{\mu_{\theta}}^s(\cdot)$ and $\boldsymbol{\sigma_{\theta}}^s(\cdot)$ are nonlinear mappings that parameterize the mean and covariance of the Gaussians, respectively, from the preceding continuous latents $z_{t-\ell}$. The mean of this probability is characterized by a simultaneously equivariant and invariant architecture. Specifically, we want the transformation of past latent $z_{t-1}$ to be preserved and conveyed to the present latent $z_{t}$ through one of the $S$ equivariant networks. However, the selection of the equivariant network is controlled invariantly by the state switch $s_t$. The diagonal covariance should, however, be characterized with an invariant network.

Finally, Gaussian distributions are defined for $p_\theta(x_{n,t}\,|\,z_{n,t})$ to map $z_{n,t}$ to the observation space $x_{n,t}$:
\begin{align}
   p_\theta(x_{t}\,|\,z_{t}) =
   \text{Norm}\big(\boldsymbol{\mu_\theta^\textbf{x}}(z_{t}),
   \boldsymbol{\sigma}^{\textbf{x}} \text{I}\big),
\end{align}
where $\boldsymbol{\mu}_{\boldsymbol{\theta}}^{\textbf{x}}(\cdot)$ is a nonlinear mapping and $\boldsymbol{\sigma}^{\textbf{x}}$ denotes the observation noise. $\boldsymbol{\mu}_{\boldsymbol{\theta}}^{\textbf{x}}(\cdot)$ is characterized by an equivariant network that preserves the transformations between input and latent (The architecture of our networks are described in Supplementary.)

\subsection{Inference Model}

The posterior distribution of our proposed model $p_\theta(\mathcal{S}, Z\,|\,X)$ is intractable. Therefore, we employ stochastic variational methods \citep{hoffman2013stochastic,ranganath2013adaptive} to learn the parameters of our model, in which the posterior of latents are approximated with a variational distribution $q_\phi(\mathcal{S},Z)$, by maximizing a lower bound on the likelihood of data, a.k.a. ELBO:
\begin{align}
    \mathcal{L}(\theta, \phi) =&\, \mathbb{E}_{q_\phi(\mathcal{S},Z)}\left[\log\frac{p_\theta(X, \mathcal{S}, Z)}{q_\phi(\mathcal{S},Z)}\right]\nonumber\\
    =&\log p_\theta(X) - \text{KL}\left(q_\phi(\mathcal{S},Z)\,\|\,p_\theta(\mathcal{S},Z|X)\right)
    \label{eqn:e}
\end{align}
The ELBO bound includes the parameters $\theta$ and $\phi$ that are related to the generative distributions, which defines the distribution over data $p_\theta(X)$, and the variational distribution, respectively. We maximize this bound with respect to the parameters $\theta$ to learn the generative model and maximize it over the parameters $\phi$ to perform inference.

We assume the following factorized variational distribution for the latents $\mathcal{S},Z$:
\begin{equation}
    q_\phi(\mathcal{S},Z) = \prod_{n=1}^N \prod_{t=1}^T
     q_\phi(s_{n,t}) q_\phi(z_{n,t}),
     \label{eqn:qinf}
\end{equation}
where $q_\phi(z_{n,t}) = \text{Norm}(\mu_\phi^{n,t}, \sigma_\phi^{n,t})$ and the categorical distributions $q_\phi(s_{n,t})$ are approximated from the posteriors $p(s_{n,t}|\tilde{z}_{n,t})$ using the Bayes' rule, where $\tilde{z}_{n,t}\sim q_\phi(z_{n,t})$, to relieve the information loss from mean-field approximation:
\begin{align}
    q_\phi(s_{t}=s)&\simeq p(s_{t}=s|\tilde{z}_{t})\nonumber\\
    &=\frac{p(s_{t}=s)p(\tilde{z}_{t}|s_{t}=s)}{\sum\limits_s p(s_{t}=s)p(\tilde{z}_{t}|s_{t}=s)}\label{eqn:vars}
\end{align}
%estimation of .. involves caluculating the distance between ... and ... which stays invariant under so3 transformation.
%\hl{explain that Gaussian probability stays invariant in SO3} 

After defining the variational structures, we insert into equation~\ref{eqn:e} the generative and variational distributions from equation~\ref{eqn:pgen} and equation~\ref{eqn:qinf}, respectively and derive the following ELBO by some algebraic manipulations. 
\begin{align}
\mathcal{L}_{t}(&\theta,\phi)=\nonumber\\
-&\mathbb{E}_{q_\phi(z_t)}\Big[\big\|x_t - \boldsymbol{\mu_\theta^\textbf{x}}(z_{t})\big\|_\text{\tiny F}^2\Big]-\nonumber\\
&\mathbb{E}_{q_\phi(s_{t-1})q_\phi(z_{t-1})}\Big[\text{KL}\big(q_\phi(s_{t})||p_\theta(s_{t}|s_{t-1},z_{t-1})\big)\Big]-\nonumber\\
&\mathbb{E}_{q_\phi(s_t)q_\phi(z_{t-\ell})} \Big[\text{KL}\big(q_\phi(z_{t})\|
     p_\theta\big(z_{t}|z_{t-\ell},s_{t})\big)\Big],\nonumber
\end{align}
where the first term corresponds to the reconstruction loss, the second term is the discrete latent loss, and the last term is the continuous latent loss. The ELBO gradients w.r.t. $\theta$ and $\phi$ are estimated using a reparameterized sample from $q_\phi(z_t)$ \citep{kingma2014auto}, i.e., $z_t=\mu_\phi^t+\sigma_\phi^t\,\epsilon$, where $\epsilon\sim \text{Norm}(0, \text{I})$, and by enumerating over the possible states in $q_\phi(s_t)$.

\subsubsection{Why do variational distributions preserve equivariance/invariance?}

While our generative design is equivariant in essence, in order to have an equivariant framework, this property needs also to be preserved in the inference design. 
The reconstruction term of ELBO $\quad\quad\quad\quad\big\|x_t - \boldsymbol{\mu_\theta^\textbf{x}}(\tilde{z}_{t})\big\|_\text{\tiny F}$, where $\tilde{z}_t\sim q_\phi(z_t)$, in conjunction with the equivariance of $\boldsymbol{\mu_\theta^\textbf{x}(\cdot)}$ encourage estimation of equivariant variational parameters (i.e., $\mu_\phi^t,\,\sigma_\phi^t$) such that the resulting posterior samples $\tilde{z}_t$ go through the same transformation as observed $x_t$. For the discrete states, estimation of their variational parameters involve computing $p(\tilde{z}_t|s_t=s)$ (see equation~\ref{eqn:vars}), which is proportional to $\exp(\|\tilde{z}_t-\boldsymbol{\mu_\theta}^s(\tilde{z}_{t-\ell})\|_{\text{\tiny F}}^2)$ (see equation~\ref{eqn:trans}), the Euclidean distance between the posterior $\tilde{z}_t$ and prior mean $\boldsymbol{\mu_\theta}^s(\tilde{z}_{t-\ell})$, which stays invariant under the SO(3) equivariance of $\boldsymbol{\mu_\theta}^s(\cdot)$

\begin{figure*}[!t]
\centering
\subfloat[a][EqDDM]{
\begin{minipage}{.24\linewidth}
\includegraphics[width =\linewidth]{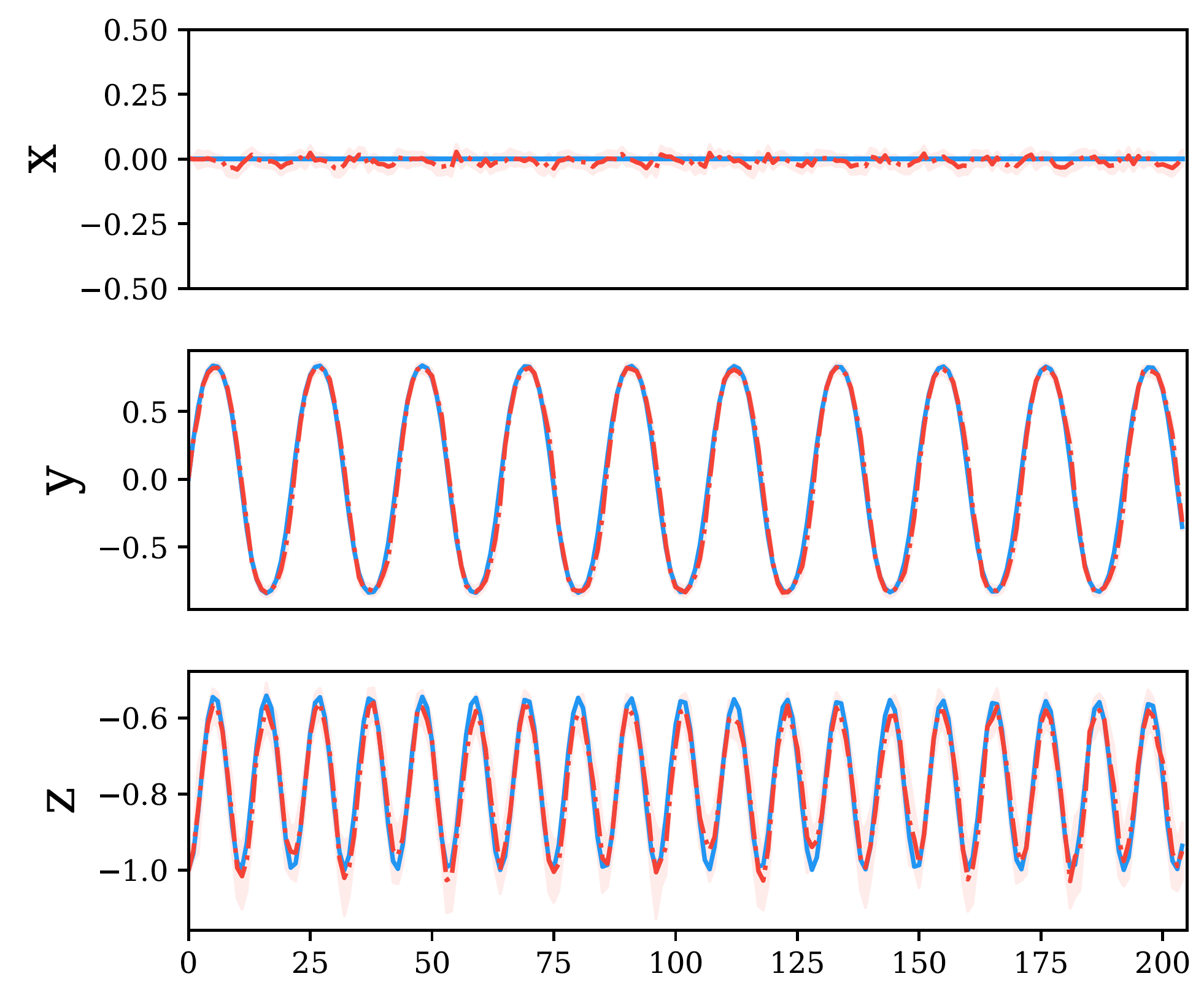}
\\
\begin{flushright}
\vspace{-.5cm}
\includegraphics[width =.9\linewidth]{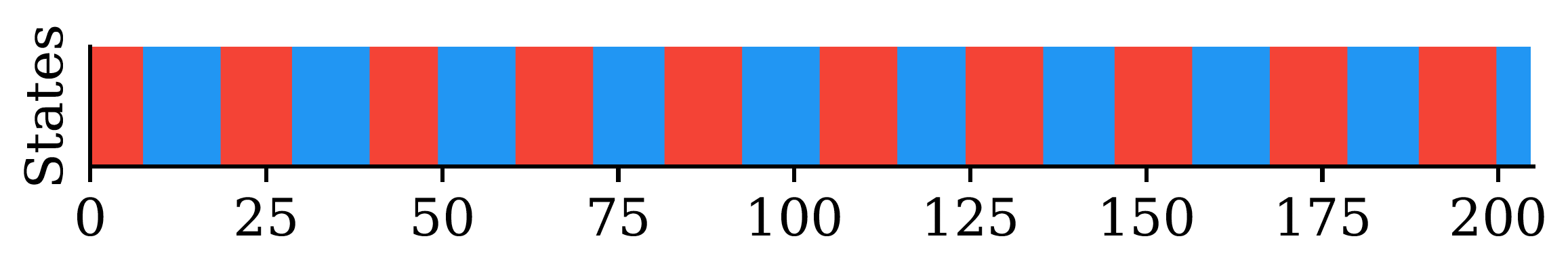}
\end{flushright}
\end{minipage}
\label{fig:eq}}
\subfloat[][DSARF]{
\begin{minipage}{.24\linewidth}
\includegraphics[width =\linewidth]{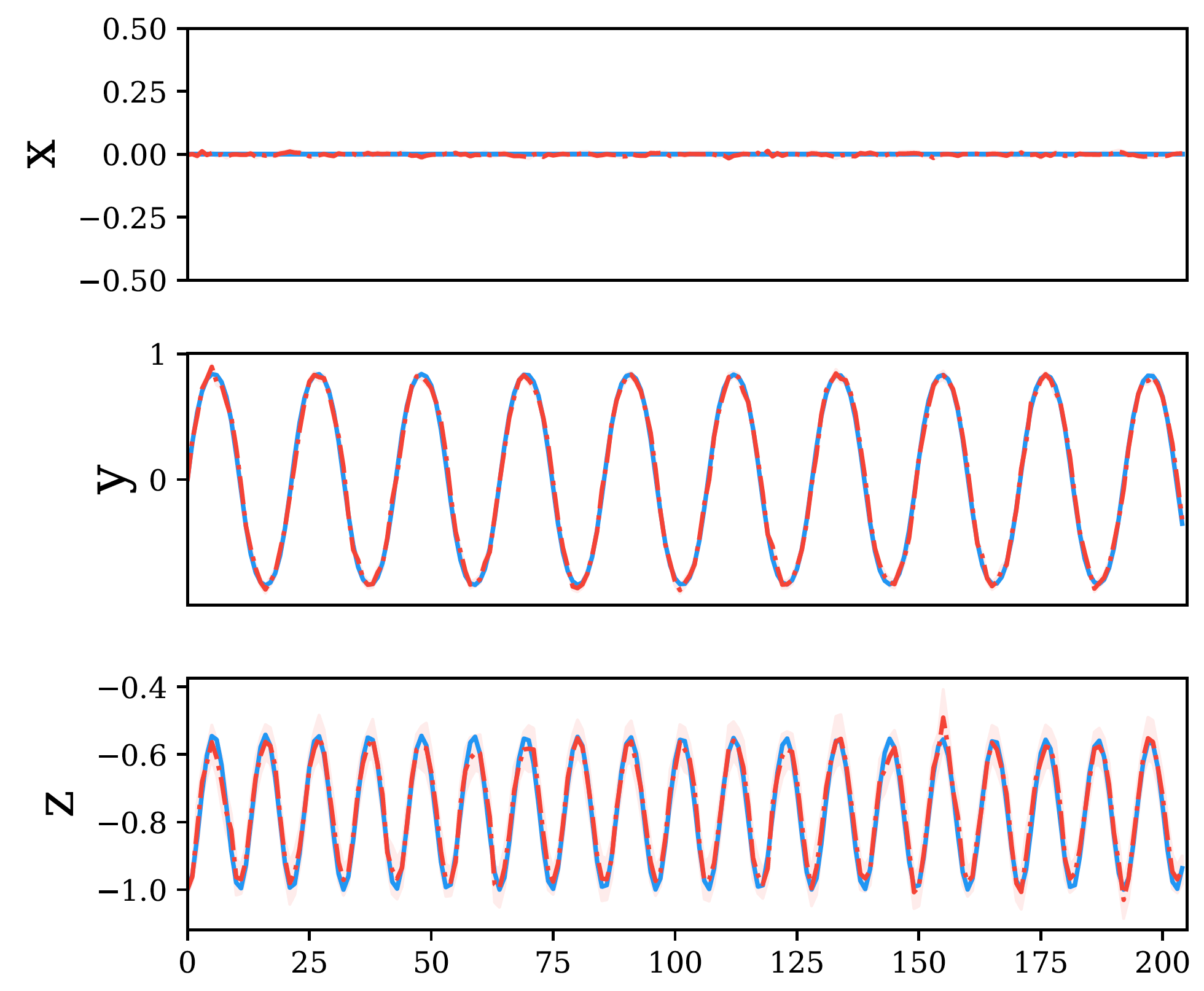}
\\
\begin{flushright}
\vspace{-.5cm}
\includegraphics[width =.9\linewidth]{fig/States.png}
\end{flushright}
\end{minipage}
\label{fig:dsarf}}
\subfloat[][EqDDM (rotated)]{
\begin{minipage}{.24\linewidth}
\includegraphics[width =\linewidth]{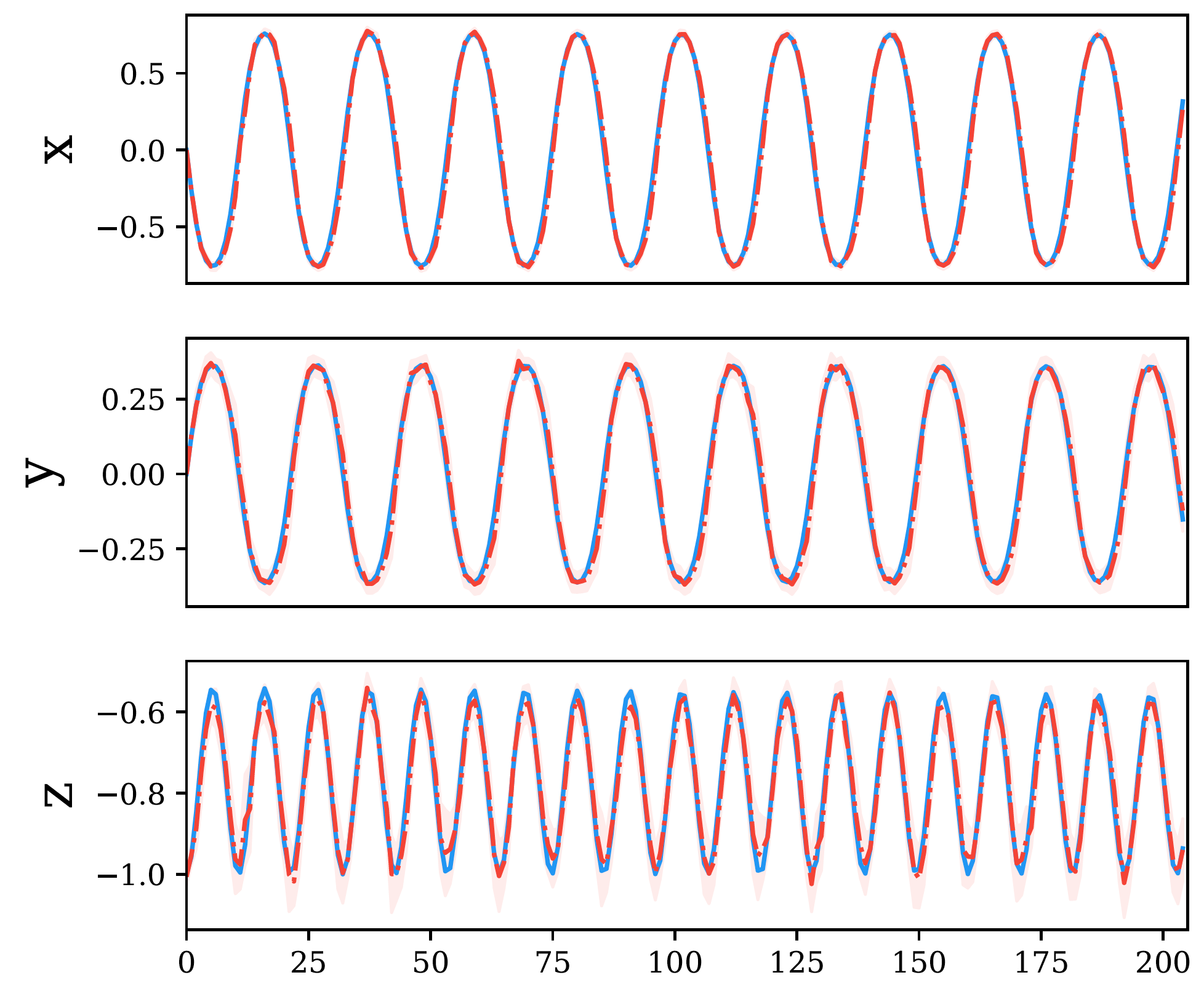}
\\
\begin{flushright}
\vspace{-.5cm}
\includegraphics[width =.9\linewidth]{fig/States.png}
\end{flushright}
\end{minipage}
\label{fig:eqrot1}}
\subfloat[][DSARF (rotated)]{
\begin{minipage}{.24\linewidth}
\includegraphics[width =\linewidth]{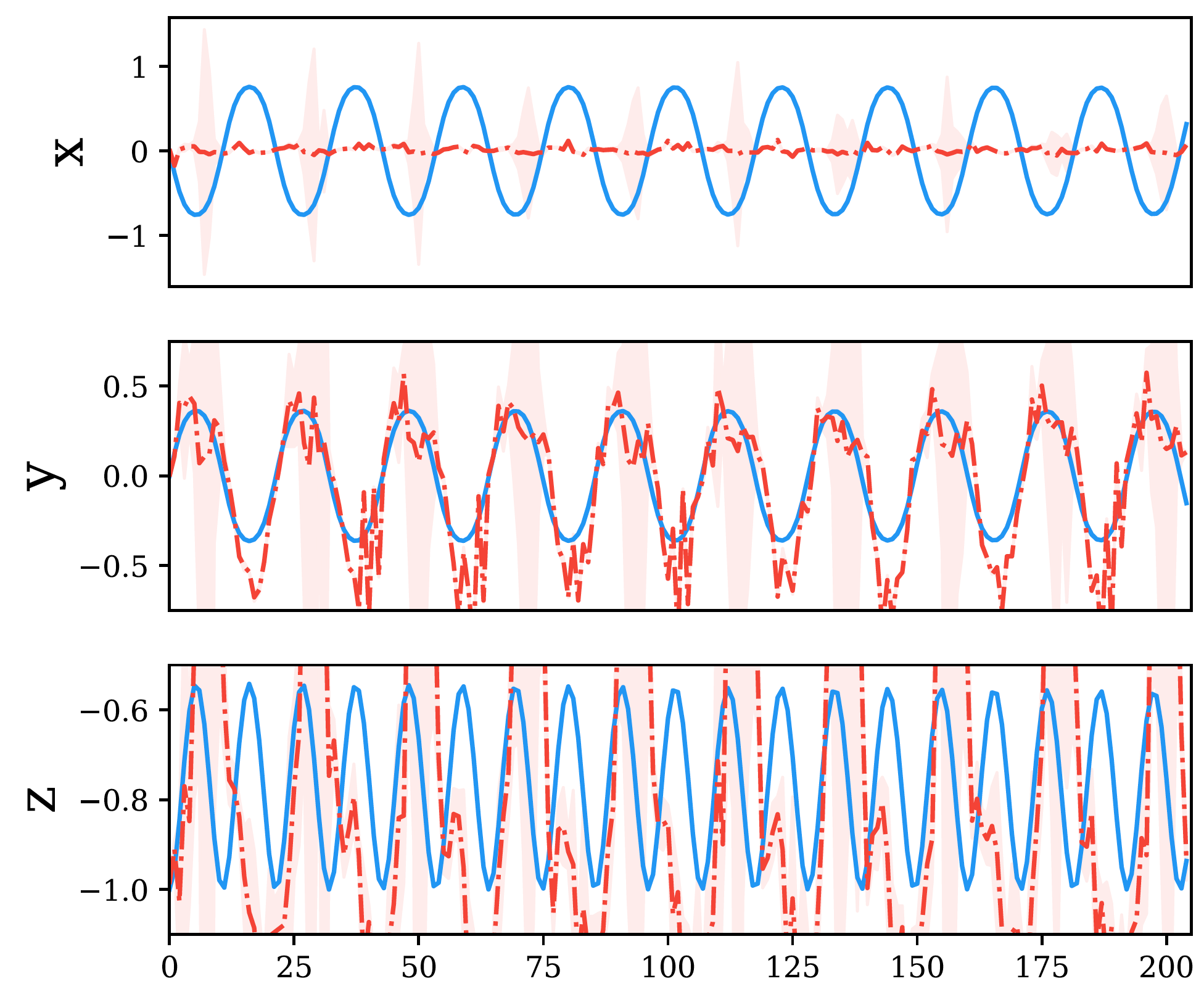}
\\
\begin{flushright}
\vspace{-.5cm}
\includegraphics[width =.9\linewidth]{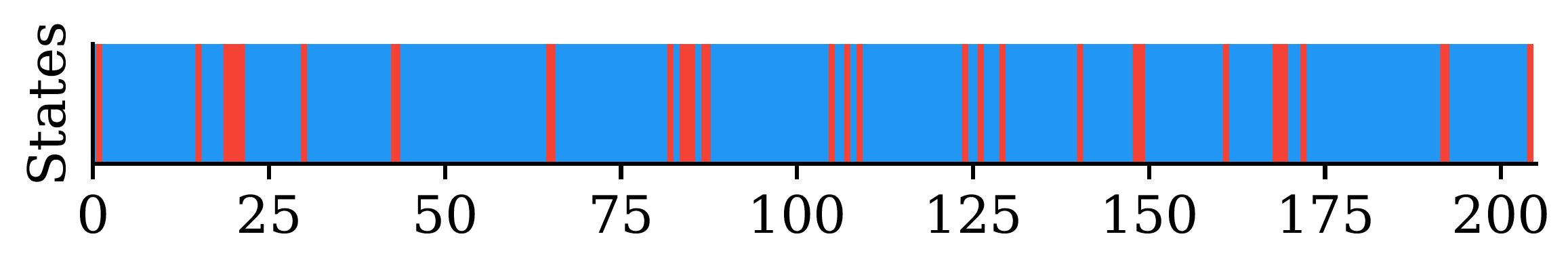}
\end{flushright}
\end{minipage}
\label{fig:dsarfrot1}}

\caption{Test set predictions (red curve) along with ground-truth (blue curve) for the pendulum experiment. The figures in the bottom row encode the inferred states: clockwise rotation (blue) and anticlockwise rotation (red). (a), (c) EqDDM successfully generalized to both the original and rotated test sets. (b), (d) DSARF performed similarly on the original test set but entirely failed on the rotated test set. This is expected as DSARF (and other baselines) are not aware of the symmetries in this dataset and overfit on the train set trajectory. As shown in the bottom row, EqDDM and DSARF decomposed the pendulum motion into two states of clockwise and anticlockwise rotation. While these states stayed unchanged for EqDDM in the rotated test set, DSARF failed to preserve its states. The red shaded regions indicate uncertainty intervals.}
\label{fig:pendulem}
\end{figure*}

\begin{figure}
    \centering
    \includegraphics[width = \linewidth]{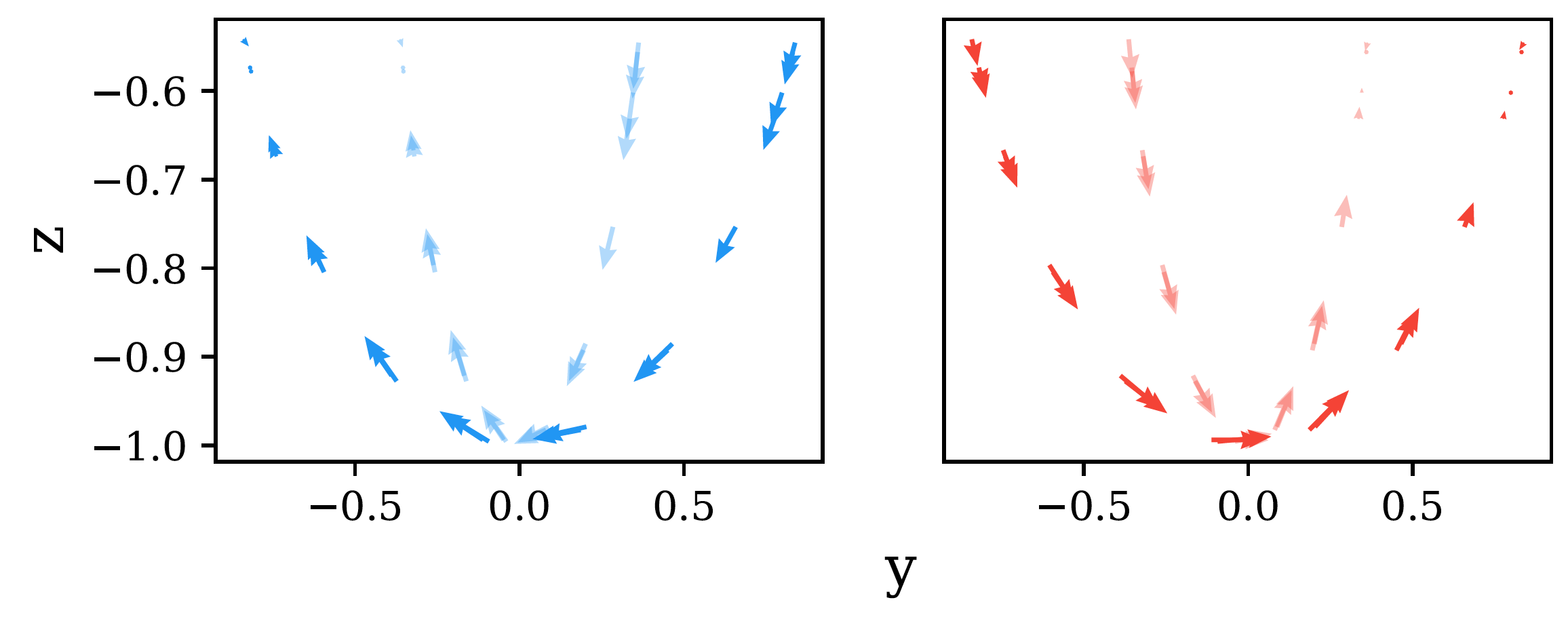}
    \caption{Dynamical trajectories of each state in the pendulum dataset (projected on the y-z plane) visualized for the original test set (solid colors) and rotated test set (shaded colors). These trajectories are computed from the generative model of EqDDM and confirm its generalizability and our interpretation of clockwise (blue) and anticlockwise (red) rotations.}
    \label{fig:my_label}
\end{figure}

\section{Experiments}
\label{sec:res}
%%%%%%%%%%%%% prediction comparison %%%%%%%%%%%
\newcommand{\ErrorTable}{
\begin{table*}[!t]
\caption{Comparison of prediction error (NRMSE\%) on the regular and rotated ($\mathcal{R}$) test sets. Our model outperforms the baselines particularly on the rotated test sets in which the baselines completely fail.}
\centering
{\small
\setlength\tabcolsep{2pt}
\begin{tabular}{lcccccc|cc}
\toprule
%\Tstrut
%\backslashbox{\scriptsize{Data}}{\scriptsize{Model}} 
\multirow{2}{*}{Dataset}
&\multirow{2}{*}{~EqDDM~}
&\multirow{2}{*}{~DSARF~}
&\multirow{2}{*}{~rSLDS~}
&\multirow{2}{*}{~SLDS~}
&\multirow{2}{*}{~~RKN~~}
&\multirow{2}{*}{LSTNet~}
&\multirow{2}{*}{~DDM+Aug.}
&\multirow{2}{*}{EqHNN}
\\
&
&
&
&
&
&
&
&
\\
\midrule
Pen.
& $5.13$
& $\mathbf{4.66}$
& $24.29$
& $27.73$
& $7.48$
& $7.33$
& $-$
& $-$
\\
$\mathcal{R}$Pen.
& $\mathbf{5.29}$
& $72.37$
& $89.88$
& $88.08$
& $78.22$
& $79.58$
& $19.37$
& $15.04$
\\
\midrule
Bat 
& $\mathbf{7.61}$
& $8.82$
& $11.39$
& $12.26$
& $19.02$
& $18.75$
& $-$
& $-$
\\
$\mathcal{R}$Bat 
& $\mathbf{7.40}$
& $47.89$
& $57.46$
& $46.93$
& $66.55$
& $69.82$
& $17.03$
& $20.82$
\\
\midrule
Golf
& $\mathbf{8.48}$
& $10.92$
& $10.60$
& $13.99$
& $12.95$
& $18.22$
& $-$
& $-$
\\
$\mathcal{R}$Golf
& $\mathbf{9.40}$
& $29.06$
& $28.73$
& $36.48$
& $32.09$
& $45.01$
& $16.98$
& $12.81$
\\
\midrule
Walk
& $\mathbf{3.86}$
& $4.72$
& $12.85$
& $13.31$
& $13.88$
& $8.70$
& $-$
& $-$
\\
$\mathcal{R}$Walk
& $\mathbf{4.53}$
& $37.21$
& $43.71$
& $41.29$
& $38.62$
& $42.94$
& $10.44$
& $8.03$
\\
\midrule
Salsa 
& $11.27$
& $\mathbf{10.94}$
& $13.25$
& $13.09$
& $13.38$
& $11.91$
& $-$
& $-$
\\
$\mathcal{R}$Salsa 
& $\mathbf{11.42}$
& $16.78$
& $18.93$
& $19.58$
& $18.97$
& $16.83$
& $15.39$
& $15.17$
\\
\bottomrule
\multicolumn{7}{l}{\scriptsize Best results are highlighted in bold fonts.}
\end{tabular}
}
\label{tbl:ErrorTable}
\end{table*}
}

\newcommand{\architecture}{
\begin{table}[!h]
\caption{Network architectures for the nonlinear mappings in EqDDM. ELL: Equivariant Linear Layer, ILL: Invariant Linear Layer, ENL: Equivariant Nonlinearity}
\centering
{
\setlength\tabcolsep{10pt}
\begin{tabular}{c|c|c|c}
\toprule
%\backslashbox{Layer}{Model} &
Network &
$\boldsymbol{\pi}_{\boldsymbol{\theta}}^s:\mathbb{R}^{K}\rightarrow\mathbb{R}^S$ 
& $\boldsymbol{\mu_{\theta}}^s,\,\boldsymbol{\sigma_{\theta}}^s:\mathbb{R}^{|\ell|\times K}\rightarrow\mathbb{R}^{K,K}$ 
& $\boldsymbol{\mu}_{\boldsymbol{\theta}}^{\textbf{x}}:\mathbb{R}^{K}\rightarrow\mathbb{R}^{3D}$
\\
\midrule
Input
& $z_{t-1}\in \mathbb{R}^K$
& $z_{t-\ell}\in \mathbb{R}^{|\ell|\times K}$
& $z_{t}\in \mathbb{R}^{K}$
\\ 
\midrule
1
& \text{ELL $K\times 3K$ ENL}
& \text{ELL $|\ell| \times K \times 5K$ ENL}
& \text{ELL $K\times 2K$ ENL}
\\
2
& \text{ELL $3K\times 3K$ ENL}
& \text{ELL $|\ell| \times 5K \times 5K$ ENL}
& \text{ELL $2K\times 2K$ ENL}
\\
3
& \text{ILL $3K\times S$ Softmax}
& \text{AvgPool($|\ell|$)}
& \text{ELL $2K\times 2K$ ENL}
\\
%\cmidrule(l{0.4em}r{0.4em}){2-2}
4
& 
& \text{ELL $5K \times 5K$ ENL}
& \text{ELL $2K\times 3D$}
\\
\multirow{2}{*}{5}
& 
& \text{ELL $5K \times K$}
& 
\\
& 
& \text{ILL $5K \times K$}
& 
\\
\bottomrule
\end{tabular}
}
\label{tbl:architecture}
\end{table}
}

\begin{figure*}[!ht]
\centering
\subfloat{
\begin{minipage}{.02\linewidth}
\rotatebox{90}{{\scriptsize Regular testset}}\\

%\vspace{-.5cm}
\rotatebox{90}{{\scriptsize Randomly rotated testset}}

\end{minipage}
\label{fig:eq1}}
\subfloat[a][bat]{
\begin{minipage}{.23\linewidth}
\includegraphics[width =\linewidth]{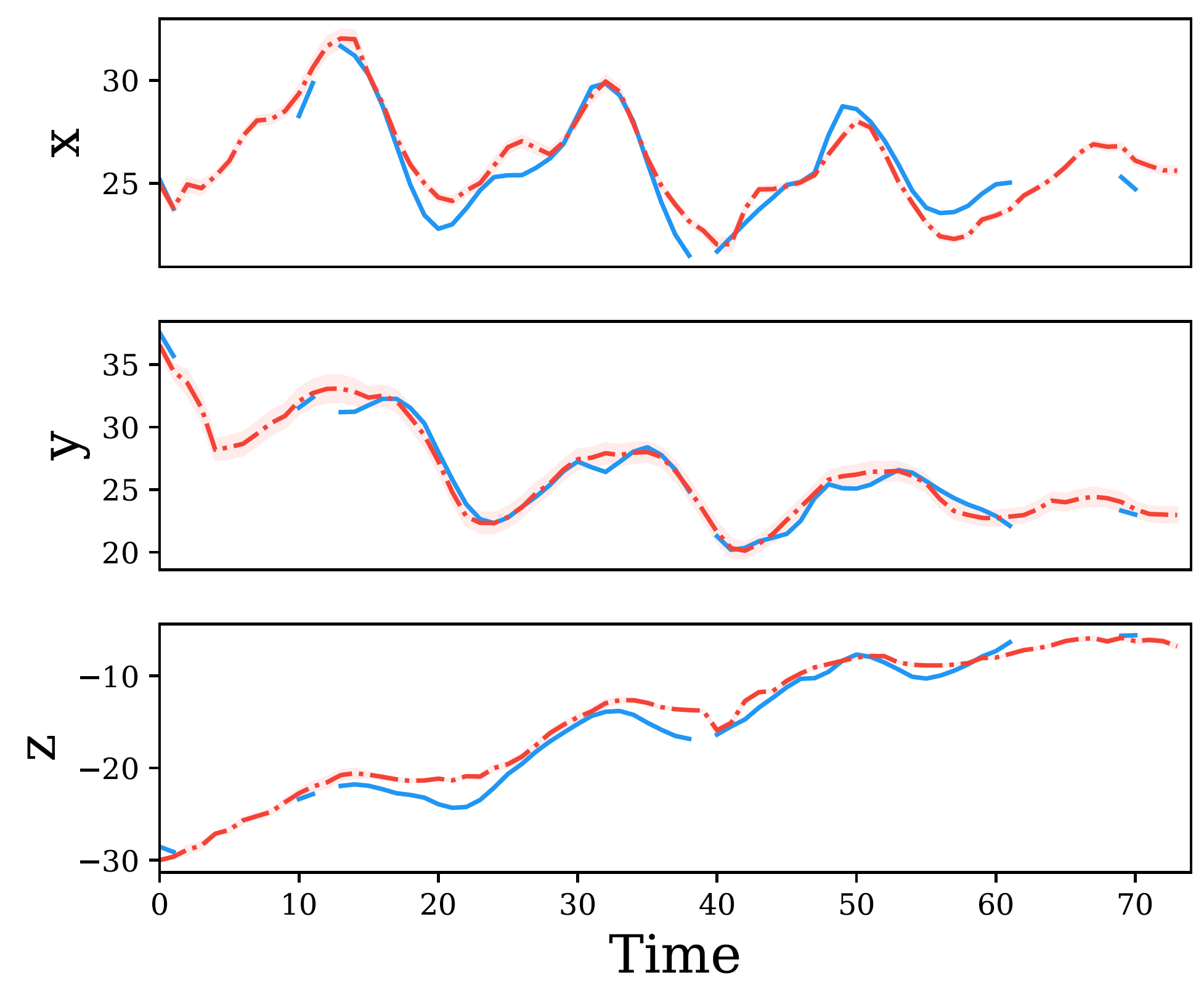}
\\
\begin{flushright}
\vspace{-.5cm}
\includegraphics[width =\linewidth]{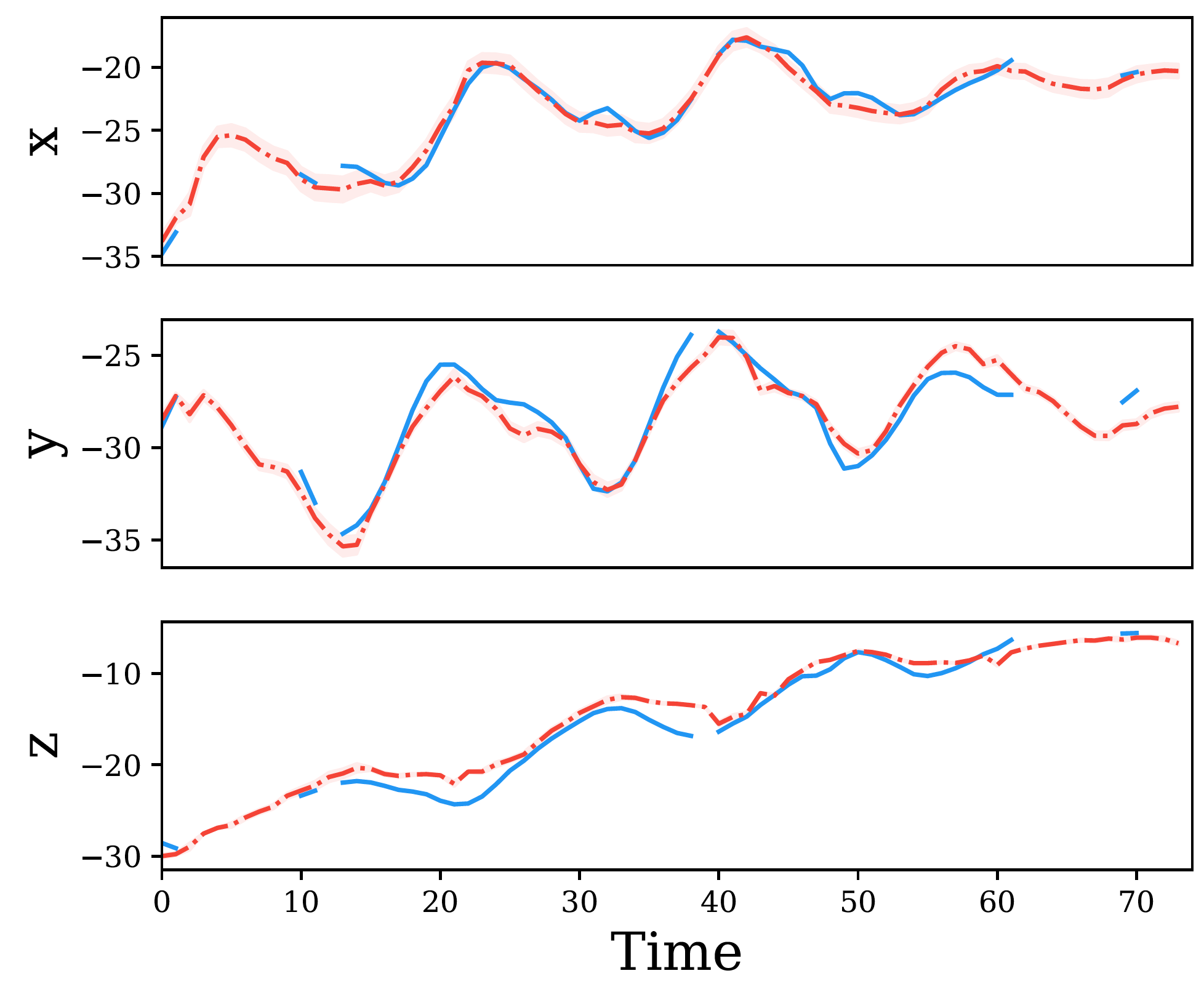}
\end{flushright}
\end{minipage}
\label{fig:eq2}}
\subfloat[][golf]{
\begin{minipage}{.23\linewidth}
\includegraphics[width =\linewidth]{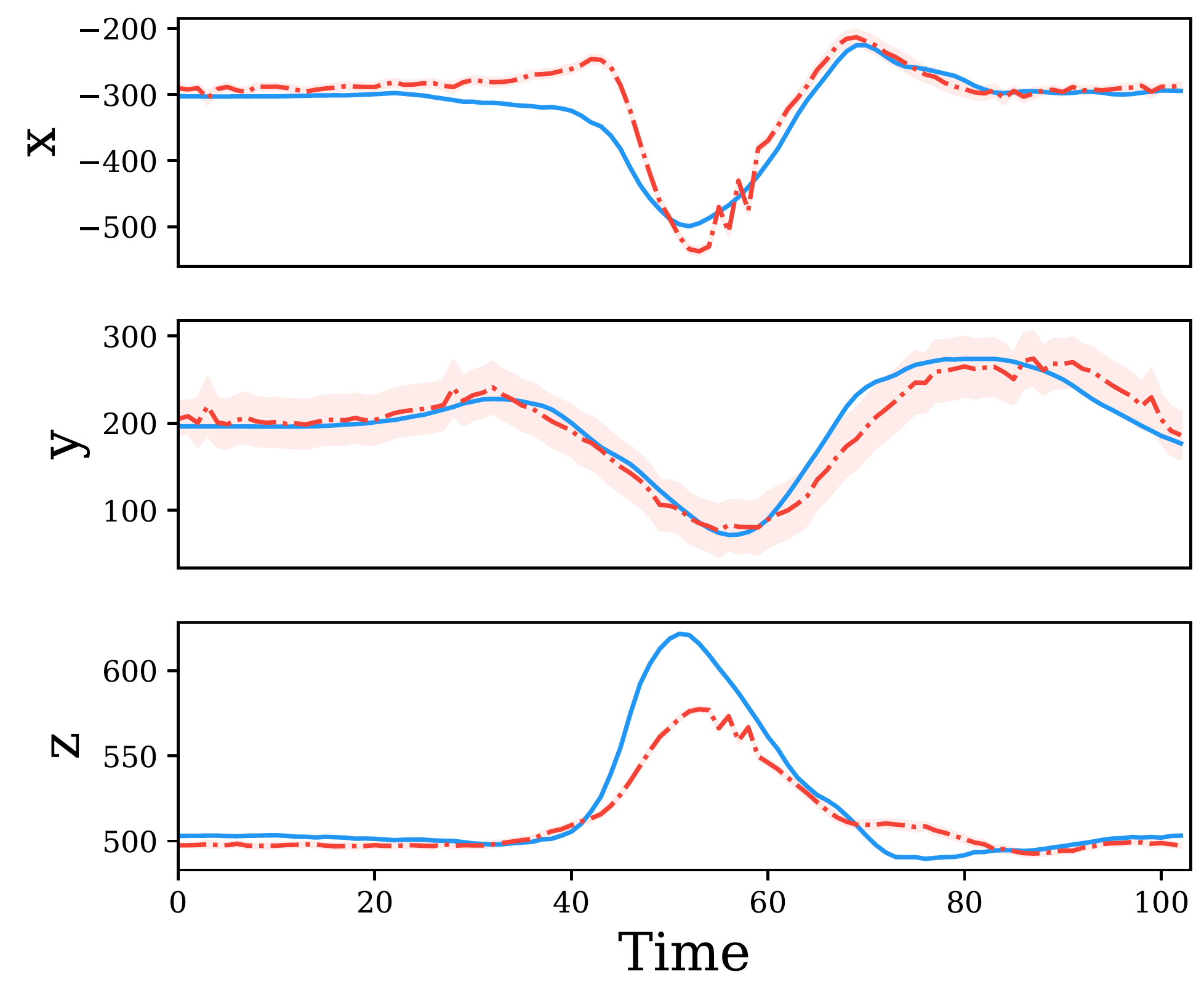}
\\
\begin{flushright}
\vspace{-.5cm}
\includegraphics[width =\linewidth]{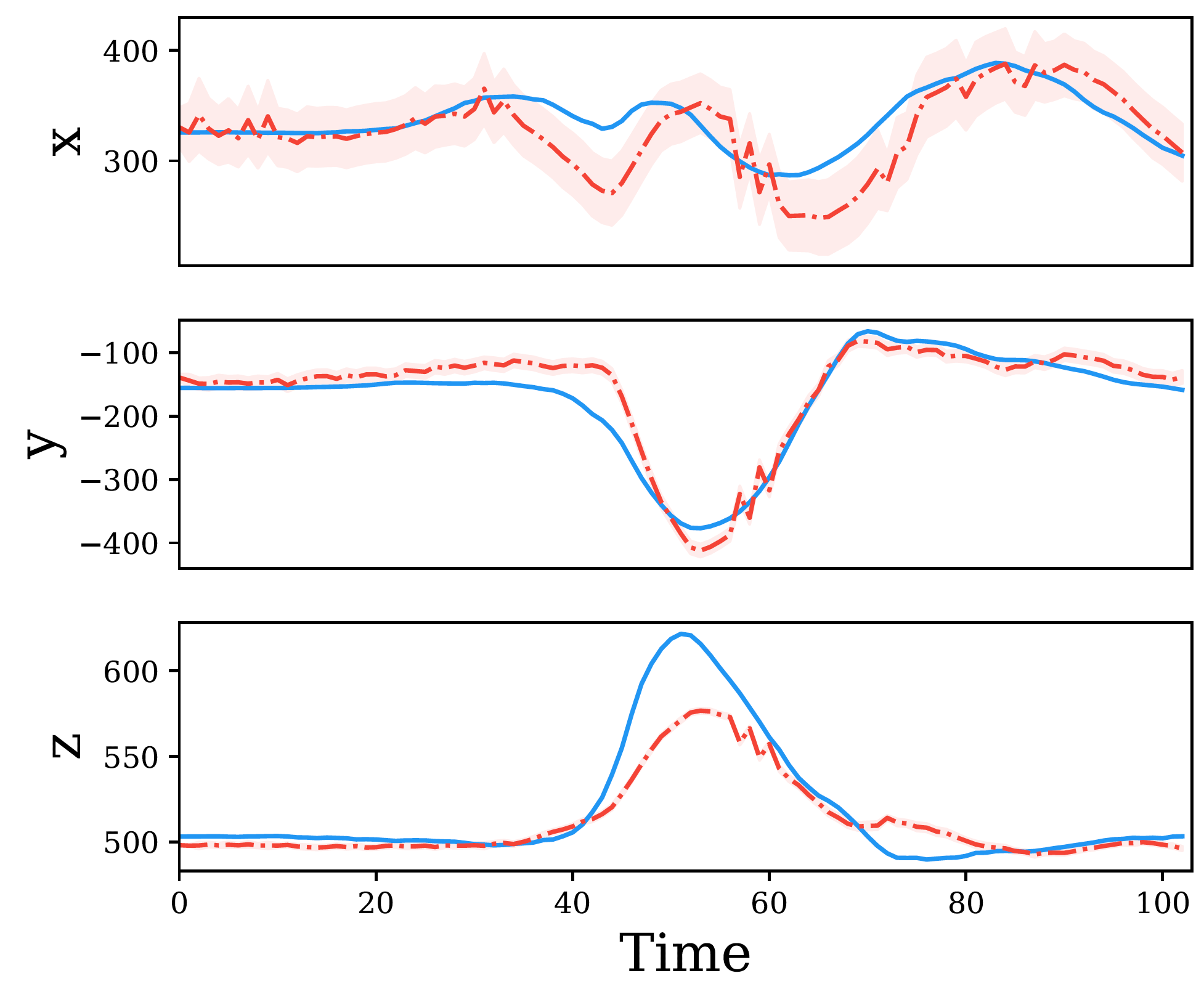}
\end{flushright}
\end{minipage}
\label{fig:dsarf1}}
\subfloat[][walk]{
\begin{minipage}{.23\linewidth}
\includegraphics[width =\linewidth]{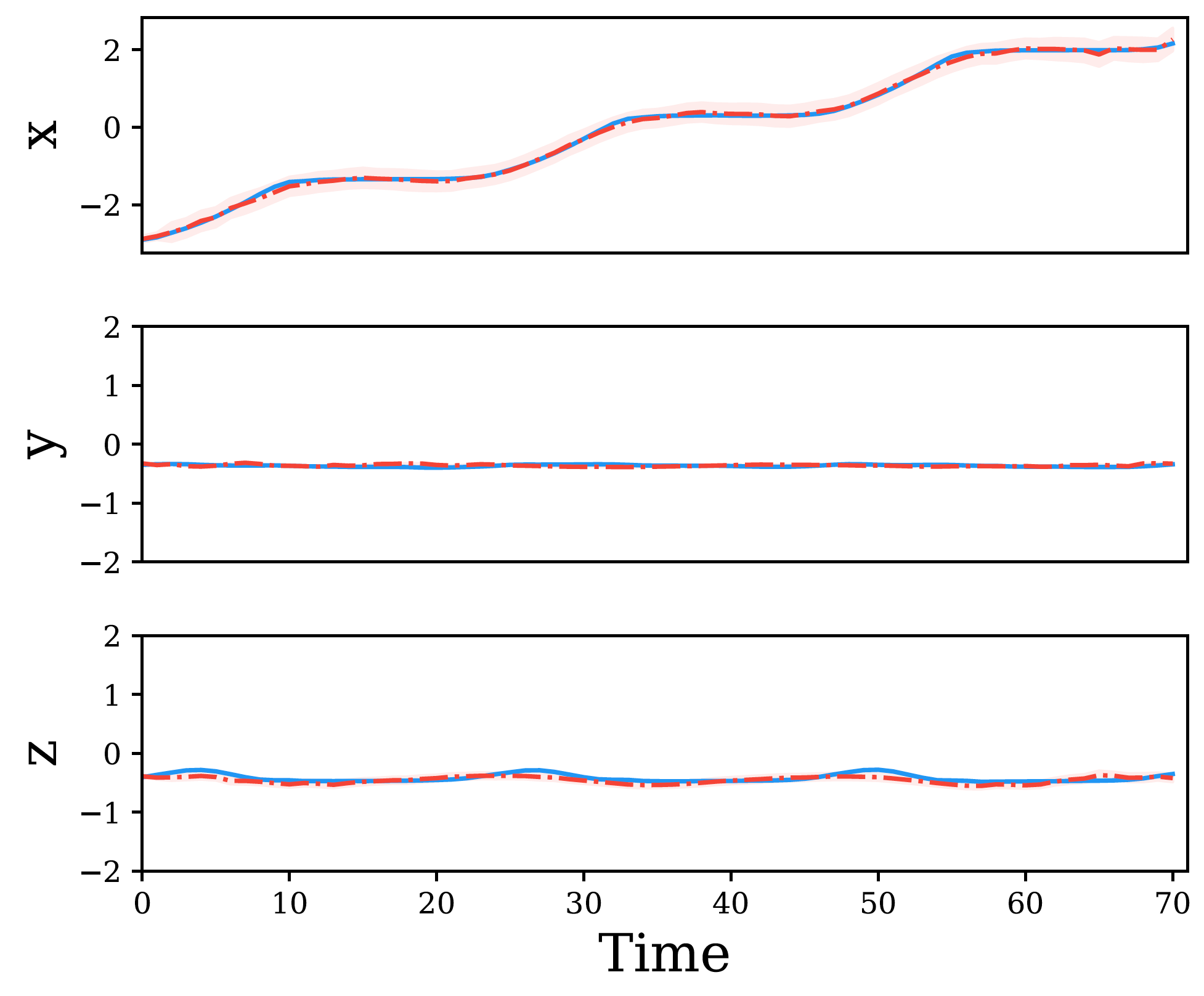}
\\
\begin{flushright}
\vspace{-.5cm}
\includegraphics[width =\linewidth]{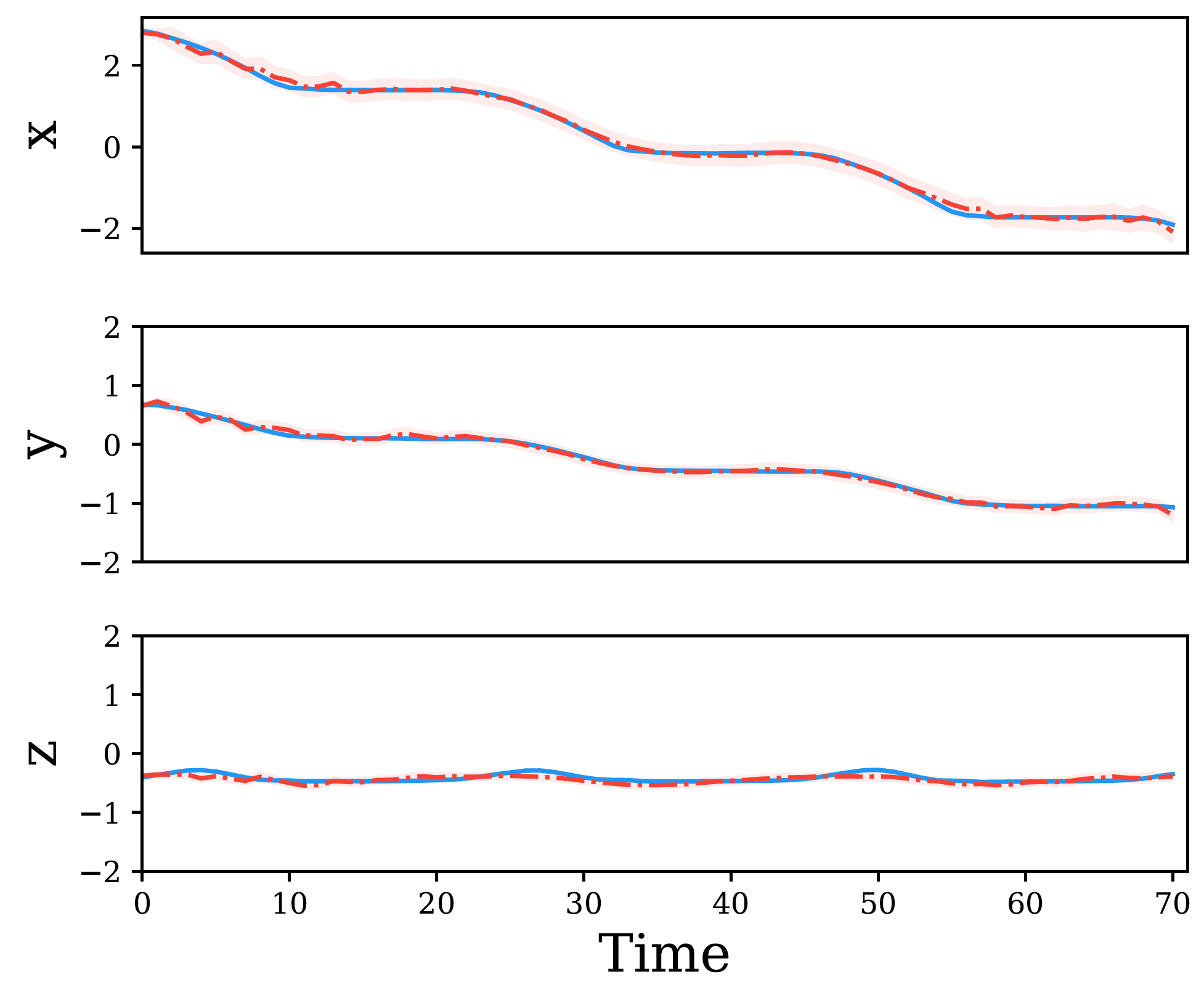}
\end{flushright}
\end{minipage}
\label{fig:eqrot}}
\subfloat[][salsa]{
\begin{minipage}{.23\linewidth}
\includegraphics[width =\linewidth]{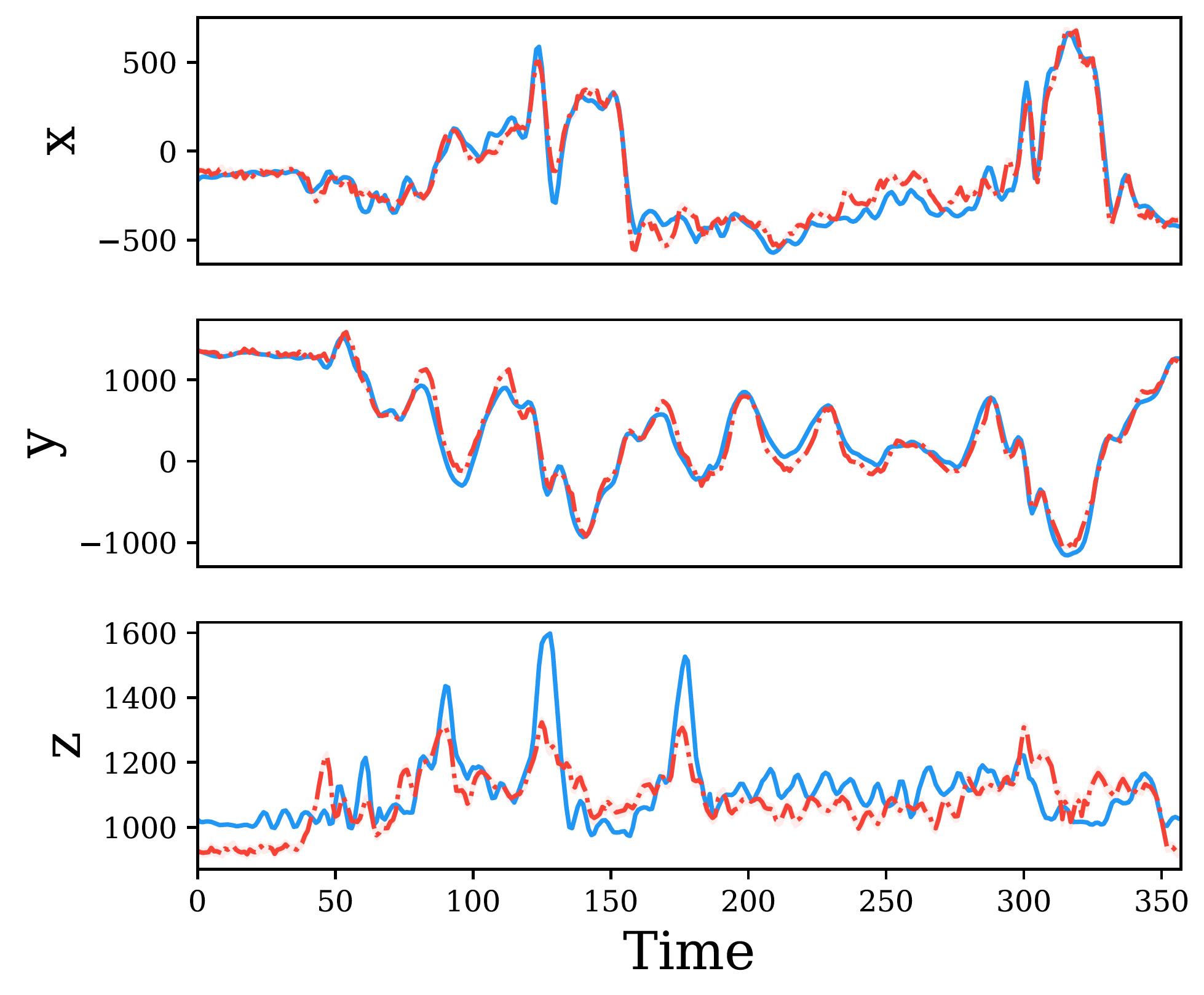}
\\
\begin{flushright}
\vspace{-.5cm}
\includegraphics[width =\linewidth]{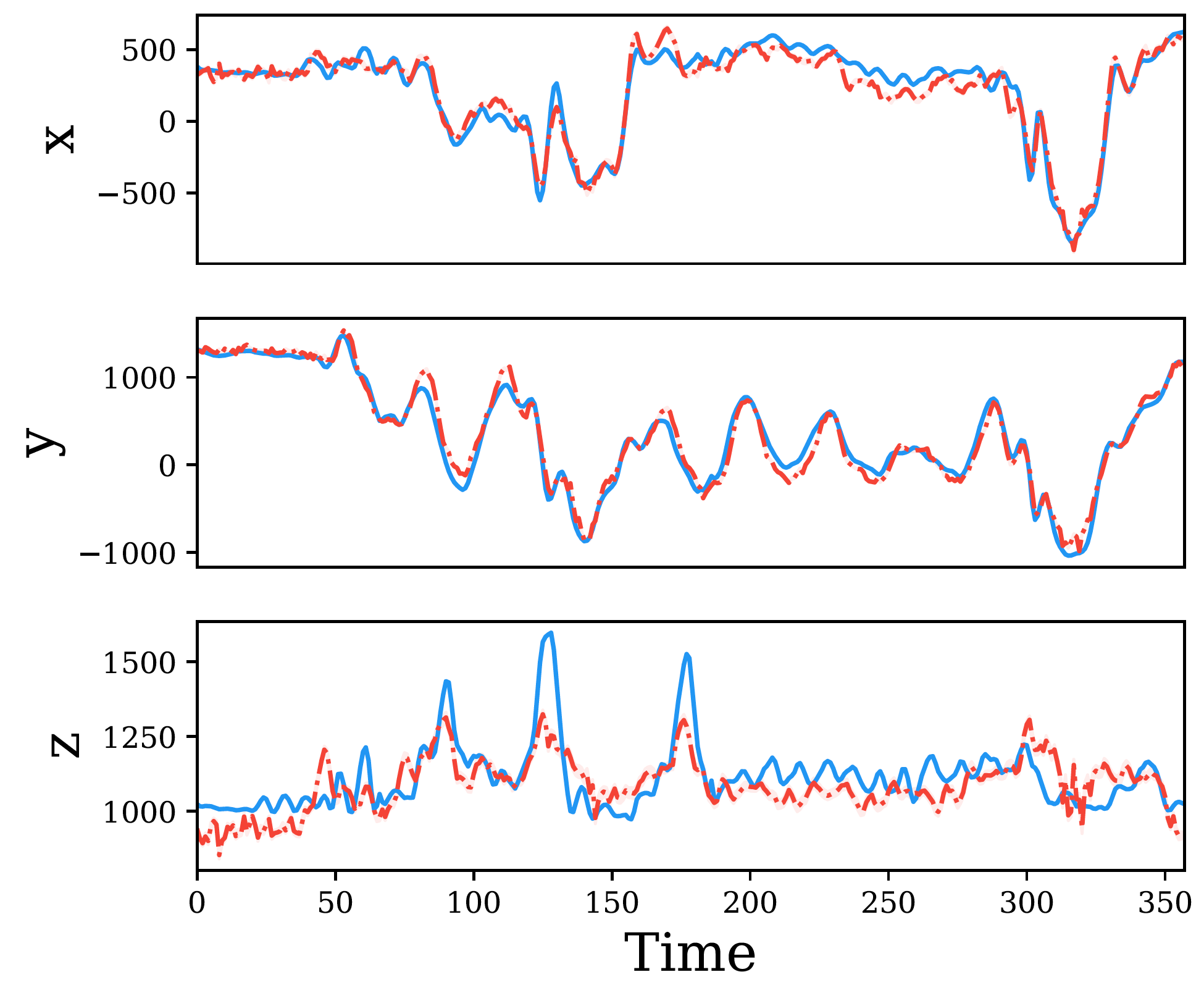}
\end{flushright}
\end{minipage}
\label{fig:dsarfrot}}

\caption{Predictions of the regular (top row) and rotated test sets (bottom row) for a sample joint from each of the real datasets. Test set predictions are shown as red curves along with ground-truth shown as blue curve.  EqDDM has successfully generalized on the rotated test sets. Also, note that our model fills in the missing values in the bat dataset. The red shaded regions indicate uncertainty intervals.}
\label{fig:4}
\end{figure*}

\subsection{Performance Assessment}
We calculate temporal predictive error in a rolling manner on a trajectory in the testing dataset to evaluate the performance of the generative model, \cite{farnoosh2021deep, linderman2017bayesian}. The next time point on the trajectory $\hat{x}_{t+1}$ is predicted using the generative model learned on the train set: $\hat{x}_{t+1}=\mu_\theta^\textbf{x}(\hat{z}_{t+1})$, where $\hat{z}_{t+1}\sim p(\hat{z}_{t+1}|{z}_{t+1-\ell}, \hat{s}_{t+1})$ and $\hat{s}_{t+1}\sim p(\hat{s}_{t+1}|s_t,z_t)$. Following that, we run inference on $x_{t+1}$, which is the true observation at $t+1$, to obtain $z_{t+1}$ and $s_{t+1}$, and add them to the historical data to predict the next time point $\hat{x}_{t+2}$. This procedure is repeated until the whole trajectory is predicted. We report the normalized root-mean-square error (NRMSE\%). Note that the generative model remains unchanged while predicting each trajectory in the testing dataset. 

%It is worthwhile to note that the test set prediction NRMSE\% is related to the expected \emph{negative test-set log-likelihood} for our case of Gaussian distributions with a multiplicative/additive constant.
\subsection{Baselines}
We compared our model with three state-of-the-art Bayesian switching dynamical models, deep switching autoregressive factorization (DSARF) \citep{farnoosh2021deep}, recurrent switching linear dynamical systems (rSLDS) \citep{nassar2019tree}, and switching linear dynamical systems (SLDS) \citep{fox2009nonparametric}, a state-of-the-art deep
state-space model, recurrent Kalman networks (RKN) \citep{becker2019recurrent}, and a deep forecasting model, long- and short-term time-series network (LSTNet) \citep{lai2018modeling}  throughout the experiments. We also compared our model with the \emph{non-equivariant} version of our EqDDM model, in which all neworks were replaced with regular MLPs, trained with augmented dataset (see DDM+Aug. in table~\ref{tbl:ErrorTable}). Finally, we implemented and compared with the equivariant version of Hamiltonian neural network of \cite{greydanus2019hamiltonian} (see EqHNN in table.~\ref{tbl:ErrorTable}). 
%%%%%%%%%%%%%%%%%%%%%%%%
%%%%%%%%%%%%%%%%%%%%%%%%%%

\subsection{Datasets}
\textbf{Pendulum:} We simulated a pendulum system on the y-z plane for $T=410$ time points and recorded its 3D coordinates. We trained the models on the first half and tested on the second half.
\textbf{Bat flight:} This dataset \citep{bergou2015falling} includes 3D coordinates of 34 joints on a bat skeleton recorded for $T$ between 33 to 87 time points (every 165 msec) during a landing/falling maneuver for 10 experimental runs with 32.55\% missing values. We kept two runs for the test.
\textbf{Golf:} This dataset from CMU MoCap\footnote{http://mocap.cs.cmu.edu/} includes $30$ trials of motion recordings from a subject while performing typical actions in a golf game. We kept two trials for the test.
\textbf{Walk:} This MoCap dataset contains 3D motion recordings from a subject for $34$ trials of walking/running. We kept two trials for the test.
\textbf{Salsa dance:} This MoCap dataset contains 3D coordinates of $19$ joints recorded for $T$ between 200 to 571 time points for 15 trials of salsa dancing. We kept one trial for the test and only used the woman dancer data.
\textbf{Rotated test sets:} Additionally, for each dataset we formed test sets by randomly rotating the original test set (about z-axis) for 10 different angles. These test sets are prefixed by $\mathcal{R}$.

\ErrorTable

\subsection{Experimental results}

\textbf{Experimental settings:} We set the number of states $S=2$, temporal lags $\ell=\{1,2\}$, the latent dimension $K=3$ for pendulum and $K=6$ for other experiments, and set the network dimensions (i.e., hidden layers) accordingly to match the number of generative parameters among comparison methods for a fair evaluation (see the Supplementary for details).

\textbf{Results:} We have summarized our experimental results in table~\ref{tbl:ErrorTable} and figures~\ref{fig:pendulem}, \ref{fig:my_label}, and \ref{fig:4}. For the pendulum experiment, as depicted in figure~\ref{fig:pendulem}, EqDDM successfully generalized to both the original and rotated test sets. EqDDM performed at par with DSARF on the original test set, however, all the baselines (including DSARF) completely failed on the rotated test set (see table~\ref{tbl:ErrorTable}). This is expected as the baselines are not aware of the symmetries in this dataset and overfit on the train set trajectory. As shown in the bottom row of figure~\ref{fig:pendulem}, EqDDM (and DSARF) decomposed the pendulum motion into two states: clockwise and anticlockwise rotation. While these states stayed unchanged for EqDDM in the rotated test set, DSARF failed to preserve its states. Figure~\ref{fig:my_label} illustrates the dynamical trajectories of each state computed from the learned generative model. This illustration confirms our interpretation of each state. For the bat, golf, and walk datasets, EqDDM consistently outperformed all the baselines and preserved its performance on the rotated test sets by exploiting the symmetries in the datasets, however, competing baselines completely failed to generalize. As reported in table~\ref{tbl:ErrorTable}, on the rotated test sets, EqDDM achieved 7.40\%, 9.40\%, and 4.53\%, respectively, while the best performing non-equivariant baseline without data augmentation only achieved 47.89\%, 28.73\%, and 37.21\%, respectively.
For the salsa dataset, EqDDM closely follows DSARF on the original test set with 11.27\% versus 10.94\%, and outperforms the other baselines. EqDDM preserves its performance on the rotated test set and surpasses all the baselines. In contrast to the other datasets in which the baselines completely failed to generalize on the rotated test set, for salsa dance these baselines achieve an acceptable performance. It is because the motions in the salsa dance are diverse enough for the models to see and memorize various rotations. However, note that EqDDM still significantly performs better than these baselines due to its inherent SO(3) equivariant design. We have visualized predictions of the regular and rotated test sets for a sample joint from each of the real datasets in figure~\ref{fig:4} which confirms the generalization capacity of EqDDM. Also, note that our model fills in the missing values in the bat dataset. As for the EqHNN model and non-equivariant version our model trained with augmented dataset (DDM+Aug.), it is intuitive to see that data augmentation helps with generalization to some extent. Nevertheless, the main shortcoming is that generalization is not guaranteed. Unlike the discrete transformation group with finite cardinality, data augmentation for the continuous groups, such as SO(3), requires introducing many transformations sampled from the continuous group.

\section{Conclusion}We proposed an SO(3) equivariant deep dynamical model for motion prediction. Our model is equipped with equivariant/invariant networks that preserve the rotational symmetry. We showcased the generalization of our model to arbitrary rotations of various motion data. %predictive performance of the proposed model on various physical systems and motion data. 

\subsubsection*{Acknowledgements}
We would like to thank Dr. Amirreza Farnoosh for providing extensive insight and expertise into this work. 

\vfill\pagebreak
\bibliography{AISTATS2022.bib}

%%%%%%%%%%%%%%%%%%%%%%%%%%%%%%%%%%%
%%%%%% SUPPLEMENT (OPTIONAL) %%%%%%
%%%%%%%%%%%%%%%%%%%%%%%%%%%%%%%%%%%

\clearpage
\appendix

\thispagestyle{empty}

% For one-column format, uncomment the following:
\onecolumn \makesupplementtitle
% For two-column format, uncomment the following:
%\twocolumn[ \makesupplementtitle ]

\section{Background on Group Theory}
In this paper, we use numerous concepts in abstract algebra, group theory, and representation theory. We provide a wider range of details on the notations and definitions we employed in the paper.

\par\textbf{Symmetry:} A symmetry is a set of The transformations should preserve the properties of the structure. Generally, it is presumed that the transformations must be invertible, i.e., for each transformation there is another transformation, called its inverse, which reverses its effect. Symmetry is thus can be stated mathematically as an operator acting on an object, are modeled by \textbf{Groups}.

\par\textbf{Group:} Let $G$ be a non-empty set with a binary operation defined as $\circ: G\times G\mapsto G$. We call the pair $(G; \circ)$ a group if it has the following properties: $G$ is closed under its binary operation (Closure), the group operation is associative –i.e., $(g_1\circ g_2)\circ g_3 = g_1\circ (g_2\circ g_3)$ for $g_1, g_2, g_3 \in G$  (Associativity axiom), there exists an identity $e \in G$ such that $g\circ e = e\circ g = g$ for all $g \in G$ (Identity axiom), every element $g \in G$ has an inverse $g^{-1} \in G$, such that $g\circ g^{-1} = g^{-1}\circ g = e$ (Inverse axiom).

\par\textbf{Subgroup:}  A non-empty subset $H$	of $G$ is called a subgroup, if $H$ is a group equipped with the same binary operation of as in $G$. We  show this as $H\leq G$. $H$ is called a proper subgroup of  if ${H}\not={G}$ and we show it as ${H}<{G}$. 

\par\textbf{Group action:} We say a group $G$ \textit{acts} on a set $\mathcal{X}$ if there exist a map $\phi: G \times \mathcal{X} \to \mathcal{X}$ such that: (1) $\phi(e,x)=x$, where $e$ is the identity element of $G$, and (ii) $\phi(g,\phi(h,x)) = \phi(gh,x)$ for all $g, h\in G$ and $x \in \mathcal{X}$. In this case, $G$ is called a transformation group, $\mathcal{X}$ is a called a $G$-set, and $\phi$ is called the group action.

\par\textbf{Lie group and infinitesimal generator:} A Lie group $G$ is a smooth manifold equipped with the structure of a group such that the group operation and inverse-assigning operation are smooth functions \citep{gilmore2006lie}. This means that the group operation and inverse operation are continuous on the manifold, and they can be expressed in terms of the coordinates. The manifold is locally represented by a chart mapping to an underlying Euclidean space $\mathbb{R}^D$, where $D$ is the dimensionality of the manifold. Furthermore, the chart map is defined in such a way that it associates the identity element in the group with the origin of Euclidean space. Elements of the Lie group can \textit{act} as a transformation on the basis an $n$-dimensional vector space known as the \textit{geometric space}, and change the coordinates of elements accordingly. We analyze Lie groups in terms of their infinitesimal generators which are the derivative of the group elements with respect to its $D$ underlying parameters at the identity. These infinitesimal generators are the basis for a new vector space, called the \emph{Lie Algebra}. 

\par\textbf{Lie algebra:} Lie algebra, denoted as $\mathfrak{g}$, is the first order infinitesimal approximation to a Lie group, and can be interpreted as a tangent space at the identity $\mathfrak{g} \coloneqq T_{\text{id}}G \subseteq \mathbb{R}^{n\times n}$. In general, an algebra over a field is a vector space equipped with a bilinear product. Thus, it consists of a set together with operations of multiplication and addition and scalar multiplication by elements of a field and satisfies the axioms of vector space and bilinear form \citep{hazewinkel2004algebras}.

\par\textbf{Lie algebra representation:} Each element of the Lie group can also be understood as a transformation on some other vector space such as what we call a Lie group of \textit{transformations} (A $N$ dimensional vector space, geometric space $G_N$ ). Every point has its own coordinate. We have some basis $e_{\mu}$, and each point in the vector space has its coordinate which will just be its component relative to the $e_{\mu}$ basis.   Each element of the Lie group represent a transformation of the basis which changes the coordinate of every point. The effect of the group elements on the elements of the underlying geometric space is describe by a function denoted as $y = f(\mu,x)$.   every Lie group is its own Lie group of transformation -- geometric space $G_N$. 
\section{Useful Tensor Manipulations}
If $A$ is an $n \times p$ matrix and $B$ an $m \times q$ matrix, the $mn \times pq$ matrix

\begin{equation}
A\otimes B=
    \begin{bmatrix}
    a_{11}B & a_{12}B & \dots & a_{1p}B\\
    a_{21}B & a_{22}B & \dots & a_{2p}B\\
    \vdots & \vdots & \ddots & \vdots\\
    a_{n1}B & a_{n2}B & \dots & a_{np}B
    \end{bmatrix}
    \nonumber
\end{equation}
is the Kronecker product of $A$ and $B$. It is also called the tensor product.
The $\textit{vec}(\cdot)$ operator creates a column vector from
a matrix A by stacking the column vectors of $A = [a_1, a_2, \dots, a_n]$
below one another as:

\begin{equation}
\textit{vec}(A)=
    \begin{bmatrix}
    a_{1}\\
    a_{2}\\
    \vdots\\
    a_{n}
    \end{bmatrix}
    \nonumber
\end{equation}
\par\textbf{Kronecker Product and the $\textit{vec}(\cdot)$ Operator:} For given matrices $A$, $B$, and  $X$ we have:

\begin{equation}
   \textit{vec}(AXB) = (B^{\top}\otimes  A) \textit{vec}(X).
\end{equation}

\section{Network architecture and experimental settings}

The network architectures for $\boldsymbol{\pi}_{\boldsymbol{\theta}}^s$, $\boldsymbol{\mu_{\theta}}^s$, $\boldsymbol{\sigma_{\theta}}^s$, and $\boldsymbol{\mu}_{\boldsymbol{\theta}}^{\textbf{x}}$ are provided in table~\ref{tbl:architecture}. For the pendulum experiment, we set the number of states $S=2$, latent dimension $K=3$, latent representation $U_\text{z}=T_1$, state representation $U_\text{s}=2T_0$, observation representation $U_\text{x} = T_1$, $\boldsymbol{\pi}_{\boldsymbol{\theta}}^s$ hidden representation $U_{h_s} = 3T_0\oplus2T_1$, $\boldsymbol{\mu_{\theta}}^s$ and $\boldsymbol{\sigma_{\theta}}^s$ hidden representation $U_{h_z} = 3T_0\oplus T_1\oplus T_2$, and $\boldsymbol{\mu}_{\boldsymbol{\theta}}^{\textbf{x}}$ hidden representation $U_{h_x} = 3T_0\oplus T_1$. For other experiments, we set the number of states $S=2$, latent dimension $K=6$, latent representation $U_\text{z}=3T_0\oplus T_1$, state representation $U_\text{s}=2T_0$, observation representation $U_\text{x} = D\,T_1$, $\boldsymbol{\pi}_{\boldsymbol{\theta}}^s$ hidden representation $U_{h_s} = 3T_0\oplus2T_1\oplus T_2$, $\boldsymbol{\mu_{\theta}}^s$ and $\boldsymbol{\sigma_{\theta}}^s$ hidden representation $U_{h_z} = 3T_0\oplus3T_1\oplus2T_2$, and $\boldsymbol{\mu}_{\boldsymbol{\theta}}^{\textbf{x}}$ hidden representation $U_{h_x} = 3T_0\oplus3T_1$. This setting of hidden representations and dimensions roughly match the generative parameter count of EqDDM with that of the baselines for a fair comparison. We set the latent dimension and number of states (if applicable) for the baselines accordingly. 

\begin{figure}
    \subfloat[$6\times30$]{
    \begin{minipage}{.33\linewidth}
    \centering
    \includegraphics[height =2.5in]{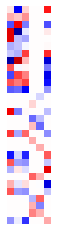}
    \end{minipage}
    }
    \subfloat[$30\times 30$]{
    \begin{minipage}{.33\linewidth}
    \centering
    \includegraphics[height =2.5in]{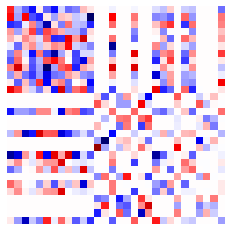}
    \end{minipage}
    }
    \subfloat[$12\times (19\times3)$]{
    \begin{minipage}{.33\linewidth}
    \centering
    \includegraphics[height =2.5in]{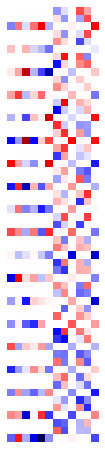}
    \end{minipage}
    }
    
    \caption{Visual representations of three equivariant linear layers for dimensions (a) $6\times 30$, (b) $30\times 30$, and (c) $12\times (19\times 3)$. For each representation, similar colors denote shared parameters.}
    \label{fig:repr}
\end{figure}

We have visualized the three equivariant linear layers with dimensions $K\times5K$, $5K\times5K$, and $2K\times 3D$, respectively in figure~\ref{fig:repr} for $K=6$ and $D=19$.

\architecture

\section{Computational Resources}
We implemented EqDDM with PyTorch v1.8 \citep{paszke2017automatic} and used the Adam optimizer \citep{kingma2014adam} with learning rate of $0.01$. We initialized all the parameters randomly. We performed all the experiments on an Intel Core i9 CPU@3.6GHz with 32 GB of RAM. Per-epoch training time varied from $200$ msec in smaller datasets to $1$ sec in larger experiments and $300$ epochs sufficed for all the experiments.

\section{Impact}
The goal of this paper is to design a deep structured architecture for generative modeling of dynamic data by adopting the formalism of group theory. We do not expect that the developed model has an immediate societal impact or poses any direct risks. However, because the problem of deep generative modeling concerns designing a probabilistic model that can generate realistic data, it may be misused in producing fake realistic data.

\end{document}